\documentclass[twoside,11pt]{article}

\usepackage[preprint]{dmlr2e}

\usepackage{booktabs}
\usepackage{amsmath}
\usepackage{microtype}
\usepackage{placeins}
\usepackage{lastpage}

\hypersetup{
  pdftitle={Lossy Compression of PDE Training Inputs: Field Reconstruction Error Does Not Order
            the Cost to a Trained Operator},
  pdfauthor={Le Huy Hoang},
  pdfkeywords={operator learning, lossy compression, rate-distortion, scientific benchmarks,
               data quality}}

\dmlrheading{1}{2026}{1-\pageref{LastPage}}{9/26}{}{26-0000}{Le Huy Hoang}
\ShortHeadings{Transmission Factors for Lossy Compression of PDE Training Data}{Le}
\firstpageno{1}

\makeatletter
\def\@maketitle{\vbox{\hsize\textwidth
 \linewidth\hsize \vskip \beforetitskip
 {\begin{center} \Large\bf \@title \par \end{center}} \vskip \aftertitskip
 {\def\and{\unskip\enspace{\rm and}\enspace}%
  \def\addr{\small\it}%
  \def\email{\hfill\small\sc}%
  \def\name{\normalsize\bf}%
  \def\AND{\@endauthor\rm\hss \vskip \interauthorskip \@startauthor}
  \@startauthor \@author \@endauthor}
  \vskip \aftermaketitskip
  \if@preprint
  \else
  \noindent \@starteditor \@editor \@endeditor
  \vskip \aftermaketitskip
  \fi
}}
\makeatother

\begin{document}

\title{Transmission Factors for Lossy Compression of PDE Training Data: Measuring What
Reaches a Trained Operator}

\author{\name Le Huy Hoang \email lhhoang@powermore.vn \\
        \addr Powermore Ltd., Da Nang, Vietnam}

\editor{}

\maketitle

\begin{abstract}%
Operator-learning benchmarks ship as full-precision arrays, and they have grown to terabyte
scale. A curator who wants to distribute one has to decide how coarsely to store it. That
decision is usually made by fixing a tolerance on the reconstruction error of the stored field.
We show that this quantity is measured in the wrong place. It compares the stored field with the
original, before any model is trained. What the curator is buying is the accuracy of a model
trained on the compressed copy, and the two can disagree: two PDEBench families stored to
identical field error differ threefold downstream. A solution operator smooths, so only part of
the codec error ever reaches the model's output. That part can be measured. Push a compressed
field through a model already trained at full precision, and compare its output against the
output on the clean field. The measurement costs forward passes and no training on compressed
data. It spans more than two orders of magnitude across six PDE families, and it orders
downstream cost where field error does not, inverting $12$ of $104$ cross-dataset comparisons
against field error's $36$. Read as a budget, the same curve nominates a storage rate for each
dataset. A back-check against trained models finds those rates wrong in both directions by up to
a factor of two in stored bits.
\end{abstract}


\section{Introduction}
\label{sec:intro}

A benchmark nobody can download is not a benchmark. PDEBench \citep{takamoto2022pdebench} and The
Well \citep{ohana2024well} ship as full-precision arrays, and they have grown to terabyte scale.
The arrays hold solutions of partial differential equations. Such fields are smooth and
repetitive, so they compress by large factors. The curator's only real decision is how far to go.
That decision is made by fixing a tolerance on the reconstructed field: store the data as coarsely
as you like, as long as the stored field stays within so many per cent of the original.

The tolerance is measured in the wrong place. It compares the stored field with the original,
before any model is trained. What the curator is buying is something else: the accuracy of a model
trained on the compressed copy. The two can disagree sharply. PDEBench stores 1D advection and 1D
Burgers on bit-identical initial conditions. At the same codec and rate the two are therefore
reconstructed to exactly the same field error. The models trained on them differ by $3.0\times$ in
cost. Over the sixteen configurations at which we trained on compressed data, field error orders
$36$ of $104$ cross-dataset comparisons backwards.

The reason is easy to state. A solution operator smooths. Part of the error the codec leaves in
the input is wiped out before it ever reaches the model's output, and how much depends on the
equation. Field error is computed before that happens, so it cannot see it.

The part that survives can be measured, and measuring it is cheap. Take a model already trained at
full precision. Push the original field through it, push the compressed field through it, and
compare the two outputs. Divide that output displacement by the input error that went in. We call
the ratio the \emph{transmission factor} $\kappa$. It costs two forward passes per rate and no
training on compressed data. Across our six families it spans more than two orders of magnitude.
Its ordering is the one each equation suggests: a shift passes everything through, an elliptic
solve almost nothing.

None of this contradicts the field-side theory, which is accurate about the field.
Rate--distortion theory applied to the covariance of a dataset gives the bits per sample needed to
hold it to a given accuracy, and Appendix~\ref{app:background} checks it on our data. It answers a
different question. Coding an observation in order to preserve something computed from it is the
indirect source coding problem \citep{wolf1970transmission}. There the two kinds of distortion
agree only when the map preserves distances, and a solution operator usually does not.

\begin{figure}[!ht]
\centering
\includegraphics{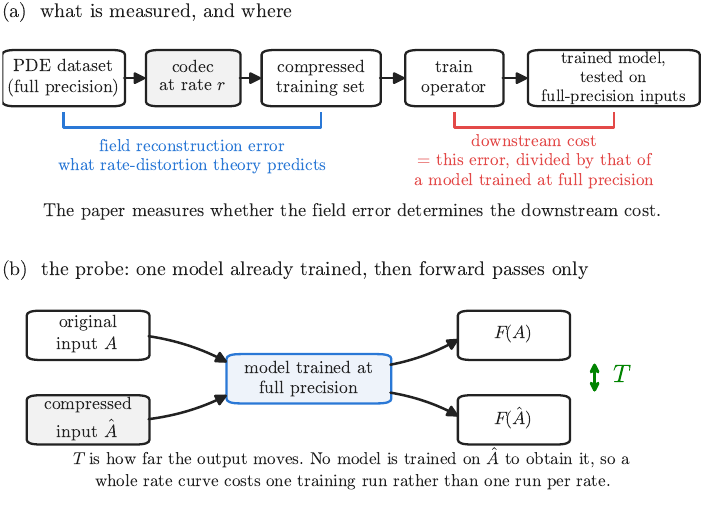}
\caption{The two quantities the paper compares. (a) Field error is measured before training and
downstream cost after it. (b) The probe replaces the training run with forward passes through
one model trained at full precision.}
\label{fig:schematic}
\end{figure}
\FloatBarrier

Our contributions are stated narrowly, in the form the measurements support.

\begin{enumerate}
\item \textbf{The transmission factor.} Measured by forward passes through one full-precision
model (Section~\ref{sec:transmit}). Its family-level medians span $\mathbf{160\times}$. Their
ordering matches how strongly each equation smooths, and it survives all three training objectives
we test.

\item \textbf{A storage rate, read off the same curve.} At a ten per cent tolerance we get
$0.1217$ bits per value on Burgers, $0.1647$ on Darcy and $0.4080$ on reaction--diffusion, which
is $263$, $194$ and $78$ times compression. On advection no rate is safe at all. Sweeping the
curve costs $2.7$ to $8.0\times$ less compute than sweeping it by training. Checked against
training, the rate is wrong in both directions by up to a factor of two in stored bits.

\item \textbf{Evidence against the field-side criterion between datasets.} Field error inverts 36
of 104 across-family cost comparisons where our estimate inverts 12, and the gap survives removing
any one family (Section~\ref{sec:decouple}). Within a single dataset field error remains reliable,
and we do not claim otherwise.

\item \textbf{What we could not do.} No mechanism survives: we tested seven explanations and
refuted all seven, and a defect in our own quantiser changed a number we had blamed on the data
(Section~\ref{sec:limitations}). We fixed 38 acceptance thresholds in advance, in the run scripts.
Fifteen were missed, among them both bars that asked the separation to survive a change of
training objective: under the reference relative-$L^2$ recipe it falls to 20 inversions against
the probe's 11.
\end{enumerate}

\section{The measurement}
\label{sec:transmit}
\label{sec:setup}
\label{sec:accounting}
\label{sec:probe}
\label{sec:rstar}

\paragraph{Data and protocol.} We use six datasets from PDEBench \citep{takamoto2022pdebench}:
advection, Burgers and reaction--diffusion in one dimension, Darcy flow and shallow water in two,
and a second reaction--diffusion set at distant parameters. Two architectures mix information
globally, a Fourier neural operator \citep{li2021fourier} and a DeepONet \citep{lu2021learning},
and one, a 1D U-Net, mixes locally. Our codec takes a discrete cosine transform and spends a bit
budget across the coefficients by reverse water-filling. Rates are quoted in bits per value (bpv),
total storage divided by the number of grid points, so single precision is $32$ bpv. Only the
stored \emph{training set} is compressed; every test error here is measured on full-precision
inputs, because a model trained on degraded inputs can learn to ignore what the codec destroyed
and look fine when tested on equally degraded data. Appendix~\ref{app:protocol} gives the
families, the splits, the architectures and the training schedule, and
Appendix~\ref{app:background} gives the allocation and the side information the decoder is charged
for, which reaches $\mathbf{3.05\times}$ the nominal rate on the smallest shallow-water settings.
Four accounting rules govern every number below, each stated because we broke it first: charge
every data-derived basis to the bit budget, hold the rate fixed rather than the distortion, state
the training objective, and check that the quantiser converged
(Appendix~\ref{app:corrections}).

\paragraph{Cost.} Write $\varepsilon(r)$ for the test error of a model trained on data stored at
rate $r$, and $\varepsilon_0$ for the same model trained at full precision. The quantity this
paper calls the downstream cost is their ratio,
\begin{equation} \mathrm{cost}(r) \;=\; \frac{\varepsilon(r)}{\varepsilon_0} . \label{eq:cost}
\end{equation}
A cost of $1.0\times$ means compression was free. A cost of $3.0\times$ means the trained model is
three times worse than it would have been. Throughout, $A$ is a full-precision input field, $\hat
A$ its compressed reconstruction, $U = F(A)$ the target, $F$ the solution operator and
$F_{\mathrm{net}}$ a trained model.

\subsection{What the operator transmits}

A model is trained on a corrupted input paired with a clean target. The codec error therefore
reaches the training objective only after the operator has acted on it. So that is where we
measure it. Take $F_{\mathrm{net}}$, trained at full precision, and feed it both versions of the
same field. The displacement at its output is
\begin{equation} T \;=\; \mathbb{E}\,\frac{\lVert F_{\mathrm{net}}(\hat A) -
F_{\mathrm{net}}(A)\rVert} {\lVert U\rVert} . \label{eq:probe} \end{equation}
Nothing here is trained on $\hat A$. $T$ is a pair of forward passes through a model that already
exists, so an entire rate curve costs one training run instead of one run per rate. Write $e(\hat
A) = \mathbb{E}\lVert \hat A - A\rVert / \lVert A\rVert$ for the field error that went in. The
ratio of the two,
\begin{equation} \kappa \;=\;
\frac{T}{e(\hat A)} \;=\; \frac{\mathbb{E}\,\lVert F_{\mathrm{net}}(\hat A) -
F_{\mathrm{net}}(A)\rVert \,/\, \lVert U\rVert} {\mathbb{E}\,\lVert \hat A - A\rVert \,/\,
\lVert A\rVert} \label{eq:kappa} \end{equation}
is the fraction of an input perturbation that survives to the output. Table~\ref{tab:transmit}
gives it for every point in this paper.

\begin{table}[!ht]
\centering
\small
\begin{tabular}{lll}
\toprule
family & $T$ / field error & the solution operator approximated \\
\midrule
advection & $0.97$ to $1.12$ & a shift, transmits everything \\
Burgers & $0.64$ to $0.93$ & nonlinear, transmits most \\
shallow water 2D & $0.35$ to $0.42$ & hyperbolic with a source \\
Darcy 2D & $0.12$ to $0.15$ & elliptic solve, smooths hard \\
reaction--diffusion & $0.032$ to $0.084$ & close to a projection \\
reaction--diffusion, far $\nu,\rho$ & $\mathbf{0.007}$ to $0.030$ & the strongest projection here \\
\bottomrule
\end{tabular}
\caption{How much of an input perturbation the trained surrogate of each solution operator
passes to its output, over all sixteen trained points.}
\label{tab:transmit}
\end{table}
\FloatBarrier

The spread is $160\times$, and it runs the way the equations do. A shift transmits everything. An
elliptic solve throws away almost all of it. One check is available. We measure $\kappa$ on the
trained model, not on the true operator $F$, and the two could differ. On advection $F$ is known
in closed form. It is a shift, which moves a field without changing its size, so the true $\kappa$
is exactly $1$. Through the surrogate we measure $0.97$ to $1.12$.

Something of the reason for the spread is visible in the output spectra
(Appendix~\ref{app:structure}). Reaction--diffusion puts $90\%$ and $99\%$ of its output energy
into a single mode, where its input needs 11 and 14. Its operator is close to a projection, so
most of the codec error is annihilated. Advection occupies the same 13 modes at input and output,
so every part of the input error passes through. Its ratio of codec error to model error is $3.21$
and its measured cost $3.03$. The spectra do not order all four families, though. Burgers' output
occupies \emph{more} modes than its input, 20 against 13, and still carries the lowest cost, and
the participation ratio reverses the direction again. We therefore offer the spectra as a
description of what differs between families, not as a mechanism.

By construction $T = \kappa\, e(\hat A)$, so everything downstream depends on the field error
through one group, $T/\varepsilon_0 = (\kappa/\varepsilon_0)\, e(\hat A)$. Two datasets stored to
the same field error are separated by $\kappa/\varepsilon_0$, and $\kappa$ alone spans
$160\times$. We also retrained all sixteen cells under two further objectives: relative $L^2$ on
normalised targets, and relative $L^2$ on the physical field, as the reference implementation
computes it. The ordering of the family-level median transmission factors did not change in
either, at a rank correlation of $1.00$ (Appendix~\ref{app:objective}).

\subsection{From transmission to a cost, and to a rate}

$T$ cannot be the cost by itself. Cost is relative to the model's own error, and a model that is
already inaccurate will not notice a perturbation that would ruin an accurate one. So the estimate
has to be a function of one dimensionless group, $x = T/\varepsilon_0$. Two limits fix its shape.
When the transmitted error is small against the model's own error, compression is invisible and
the cost is $1$. When it is large, the model reproduces the damage and the cost grows with it. The
simplest smooth function with $f(0) = 1$ and $f(x) \to x$ is
\begin{equation} \widehat{\mathrm{cost}} \;=\;
f\!\left(\frac{T}{\varepsilon_0}\right) , \qquad f(x) \;=\; \sqrt{1 + x^2} , \label{eq:pred}
\end{equation}
which is also what two independent error sources would give. We call this estimate the
\emph{probe}. The two limits are properties of the regime; the curve between them is a choice.
Section~\ref{sec:limitations} reports that the independence reading fails its own test, so the
probe is empirical, with a measured error bar and no derivation behind it.

A cost estimate at one rate is still not what a curator wants. The curator wants the rate itself:
the smallest budget that keeps the model acceptable. Fix a tolerance $t$, say $10\%$. Then
$\widehat{\mathrm{cost}} < t$ holds exactly when $T/\varepsilon_0 < \sqrt{t^2-1}$, so
\begin{equation} r^\star(t) \;=\;
\min\bigl\{\, r \;:\; T(r) \;\le\; \sqrt{t^2-1}\;\varepsilon_0 \,\bigr\} , \label{eq:rstar}
\end{equation}
is where the transmission curve crosses a fixed multiple of the model's own error. At $t = 1.10$
that multiple is $0.458$. The curve costs forward passes only, so it can be swept densely for the
price of one training run. Appendix~\ref{app:background} gives the field error this allows.

\begin{table}[!ht]
\centering
\begin{tabular}{llrrr}
\toprule
family & dim & fp32 error & $r^\star(1.10)$ & compression \\
\midrule
Burgers            & 1 & $9.275\%$ & $0.1217$ bpv & $\mathbf{263\times}$ \\
Darcy              & 2 & $4.215\%$ & $0.1647$ bpv & $\mathbf{194\times}$ \\
reaction--diffusion & 1 & $0.097\%$ & $0.4080$ bpv & $\mathbf{78\times}$ \\
advection          & 1 & $0.857\%$ & $> 8$ bpv     & none at $10\%$ \\
\bottomrule
\end{tabular}
\caption{The candidate rate nominated by the probe, the smallest at which the trained operator
may stay within $10\%$ of its full-precision accuracy. Advection never reaches it: its curve
flattens at $1.11\times$, where the $14$-bit cap leaves $0.36\%$ field error at any rate.}
\label{tab:rstar}
\end{table}
\FloatBarrier

Checked against training, $r^\star$ inherits the probe's magnitude error, in both directions. On
Darcy and on the far-parameter reaction--diffusion set, a trained rate above the nominated
$r^\star$ costs $1.27\times$ and $1.15\times$, outside the $10\%$ budget that defines it. On
reaction--diffusion a trained rate $1.6\times$ below it already costs only $1.05\times$. On
Burgers the trained points agree. Fitting the trained points instead of the probe curve moves
$r^\star$ by $1.17$ to $2.04\times$ in stored bits. So $r^\star$ is a candidate rate to validate
by training, and the 4 significant figures in Table~\ref{tab:rstar} are not warranted. We keep the
table because the spread across families is the point, and it is large
(Appendix~\ref{app:rstarcheck}). Sweeping an eight-rate curve by training costs $2.69$ GPU-hours
per family on our hardware. The probe curve costs $0.34$, or $1.01$ with two rates validated by
training, so the saving is between $2.7$ and $8.0\times$ (Appendix~\ref{app:cost}).

\section{Evidence}
\label{sec:decouple}

Start with the cleanest case. PDEBench stores two of our families on identical initial conditions,
so at one codec and one rate they are reconstructed to one field error.

\begin{table}[!ht]
\centering
\begin{tabular}{lrrrr}
\toprule
family & field error & fp32 error & cost & seeds \\
\midrule
advection          & $2.743\%$ & $0.855 \pm 0.010\%$ & $\mathbf{3.03 \pm 0.21\times}$ & $5$ \\
Burgers            & $2.743\%$ & $9.275 \pm 0.022\%$ & $\mathbf{1.01 \pm 0.00\times}$ & $2$ \\
reaction--diffusion & $1.807\%$ & $0.097 \pm 0.000\%$ & $\mathbf{1.05 \pm 0.00\times}$ & $2$ \\
\bottomrule
\end{tabular}
\caption{Same codec and rate ($0.25$ bits per value, $128\times$ compression), corrected
codebook. The first two rows share their inputs bit for bit, and so their field error, and
differ by $3.0\times$ downstream.}
\label{tab:decouple}
\end{table}

The pattern is not special to that rate or those two families. Sixteen measurements across six
families span field errors from $0.4\%$ to $35\%$. Figure~\ref{fig:headtohead} plots them all, and
Table~\ref{tab:decouple8} in Appendix~\ref{app:blocks} lists them.

We score a predictor by counting the pairs it gets backwards. Take a candidate predictor $q$ and a
set $P$ of measured points, each with a measured cost $c$. Then
\begin{equation} I(q; P) \;=\;
\bigl|\bigl\{ \{x,y\} \subset P \;:\; \bigl(q(x) - q(y)\bigr)\bigl(c(x) - c(y)\bigr) < 0
\bigr\}\bigr| , \label{eq:inv} \end{equation}
counts the pairs where the predictor and the cost disagree about which point is worse. Pairs
within one family and pairs across two families answer different questions, so we count them
separately. Within a family, both $q$ and $c$ fall as the rate rises, so every predictor in this
paper scores $I = 0$ of 16. That column is a property of the sweep, not a result. The
across-family pairs are the ones that discriminate, and there field error gives $I = 36$ of 104
under mean-squared-error training.

\begin{figure}[!ht]
\centering
\includegraphics{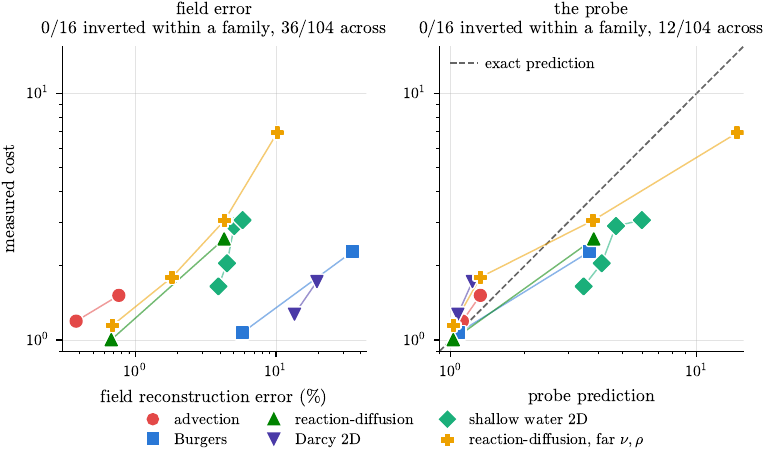}
\caption{The sixteen trained measurements of Table~\ref{tab:decouple8}: measured cost against
field error on the left and against the probe on the right, points of each family joined.
Corrected codebook, mean-squared-error training.}
\label{fig:headtohead}
\end{figure}

Judged on the same data, field error orders $36$ of the $104$ across-family pairs the wrong way,
while the probe orders $12$. Neither inverts a single within-family pair.

Those 104 pairs come from 15 family-pair blocks, so they are not independent. We therefore
resample whole blocks. The block bootstrap separates the two predictors cleanly, at $0.12$ $[0.07,
0.16]$ against $0.35$ $[0.20, 0.51]$, with disjoint intervals. It puts field error's inversion
rate at $0.35$, $95\%$ interval $[0.20, 0.51]$. Count a block as lost when a predictor gets more
than half of its pairs backwards. \textbf{The probe loses none of the 15 family pairs} and field
error loses four.

The gap also survives dropping families. We removed each family in turn, then the pair that shares
initial conditions, then the two added last. Field error inverts $26$ to $41\%$ of the remaining
across-family pairs and the probe $11$ to $13\%$, and the probe loses no block in any subset
(Appendix~\ref{app:blocks}). A family-level permutation test gives the probe $p = 0.047$. Six
families give that test little resolution, so the case rests on the size of the gap and on its
stability, not on the test.

Field error is not noise. Its sign is simply unreliable. It is exact on three of the fifteen
family pairs and gets $6$ or $7$ of $8$ backwards on two others, losing $4$ blocks, tying $3$ and
winning $8$. Nor is the failure an artefact of the seven rates we trained at. The probe curve can
be swept over every rate the six families share, from $0.0625$ to $4$ bits per value. At each of
them field error inverts between $4$ and $9$ of the 15 family pairs, against the $7.5$ that chance
would give.

\paragraph{Which ingredient carries the ordering.} Cost is a ratio to $\varepsilon_0$, and so is
the probe. Field error knows nothing about $\varepsilon_0$, which spans $0.021\%$ to $9.3\%$ across
our families. Normalising field error by $\varepsilon_0$ does help, from $36$ inversions to $29$.
And $\varepsilon_0$ on its own, which carries no information about the codec at all, does about as
well as field error, at $38$ (Table~\ref{tab:ablation}). So part of what we have attributed to
field error is really its blindness to a free family-level quantity. Neither ingredient reaches
the probe's $12$, and the transmitted error $T$ alone is at chance. \emph{Only the ratio orders the
cost.}

\paragraph{Where the probe itself fails.} Its ordering is reliable; its magnitude is not. Every
figure quoted with a $\times$ here is the symmetric ratio
\begin{equation}
\rho \;=\; \max\!\left\{ \frac{\widehat{\mathrm{cost}}}{\mathrm{cost}} ,\;
                          \frac{\mathrm{cost}}{\widehat{\mathrm{cost}}} \right\} \;\ge\; 1 ,
\label{eq:rho}
\end{equation}
which bounds the error in either direction. Inside our own codec, over 48 measured points, the
worst was $\mathbf{1.61\times}$ on the first four families, with a median of $1.14\times$. It rose
to $2.10\times$ when a fifth family arrived, and to $\mathbf{2.85\times}$ with the U-Net. Across
codecs the worst is $\mathbf{4.77\times}$. Inside a Daubechies wavelet, built by changing the
transform and nothing else, two bars taken from our own figures were missed at $3.49\times$ and
$2.14\times$. A wavelet is a transform codec like ours, so the accurate case is the particular
codec we built, and not a class of codec. We therefore stop quoting a single calibration number
and present the probe as a way to rank (Appendix~\ref{app:probe}).

\paragraph{Scope of the claim.} Field error is not useless. Compare five codec arms at a fixed
budget inside one family, and it orders all 10 arm pairs correctly on reaction--diffusion, though
it inverts 3 of 10 on advection. Our claim is the narrow one. \emph{Across} families it does not
order the cost reliably, and how badly it fails depends on the training objective. Under the
reference relative-$L^2$ recipe it falls to 20 inversions against the probe's 11, which misses a
bar we had pre-registered (Appendix~\ref{app:objective}).

\section{Two further uses}
\label{sec:ratefair}
\label{sec:frontier}

The same accounting answers two questions a curator asks next, and
Appendices~\ref{app:ratefair} and \ref{app:frontier} work both out in full.

\emph{Which codec, and charged how.} SZ3 \citep{liang2022sz3} and ZFP
\citep{lindstrom2014fixed} store each field on its own, and each stored field carries a header, an
exponent and at least one coded plane. That fixed cost puts a floor under the rate: $0.898$ bits
per value for SZ3 and $0.375$ for ZFP on a $1024$-point field, against the $0.0625$ this paper
operates at. \textbf{At our operating point neither compressor can deliver an independently
decodable field at all.} There is no single winner across rates either: our spectral code beats
real SZ3 by $45.6\times$ on advection at a quarter of a bit, while SZ3 leads by more than two
orders of magnitude at 8 bits. One family needs no approximation at all, since PDEBench draws the
Darcy permeability field from two values that SZ3 stores \emph{losslessly} at $0.404$ bits per
value.

\emph{Samples or precision.} Ten thousand fields at a quarter of a bit and two hundred and fifty
at full precision cost the same. Which trains the better model? We walked four budget lines and
registered in advance that the best point would sit at the largest sample count. \textbf{That
expectation failed on 3 of the 4 lines.} On advection at a quarter of a bit the best point stores
$250$ fields at $7.73$ bits per value and beats the full pool by $\mathbf{2.06\times}$, and the
full pool is the \emph{worst} point on that line. On reaction--diffusion the best point sits at
the largest or next-largest sample count. These are the two families that transmit almost all and
almost none of their codec error, so $\kappa$ also says which way to spend a budget. Two families
cannot establish that, and we report it as the reading of four lines.

\section{Related work}
\label{sec:related}

\paragraph{How much data an operator needs.} \citet{lanthaler2024operator} counts bits of the
\emph{model} and says so: the focus there is parametric rather than data complexity.
\citet{kovachki2024data} give data complexity estimates, \citet{adcock2024sample} sample
complexity for Lipschitz operators under Gaussian measures, and \citet{adcock2025sharp} minimax
risk bounds. That literature counts \emph{how many} samples an operator needs. This paper counts
how many \emph{bits each sample} needs, and Section~\ref{sec:frontier} shows the two trade against
each other at fixed storage. \citet{luschgy2004sharp} give the sharp quantisation rates the
field-side framework implies, and \citet{kovachki2023neural} the framework these architectures sit
in.

\paragraph{Compressing scientific data, and judging the result.} SZ3 \citep{liang2022sz3} and ZFP
\citep{lindstrom2014fixed} are error-bounded compressors, task-agnostic by design. MGARD
\citep{ainsworth2019mgard} exists partly because that is a limitation: it supplies computable
bounds on quantities of interest. Learned compressors replace the prediction stage with a network
while keeping the error bound \citep{liu2023srnsz}. That point-wise field error is an inadequate
acceptance test is roughly a decade old in that community. \citet{baker2016evaluating} built an
ensemble-based test for climate model output precisely because field norms did not settle whether
the diagnostics survived, and concluded that lossy compression cuts data volumes with minimal
effect on scientific results. Closest of all, \citet{underwood2024effectiveness} asked our question
directly, over seventeen reduction methods and seven machine-learning applications, and reached
the same permissive answer. Our result does not contradict theirs. A well-chosen tolerance is
usually safe inside one application; across datasets it cannot be chosen from the field alone.
Closest in aim, \citet{li2026quantifying} ask how much compression error a neural \emph{generative
surrogate} tolerates, setting a tolerance per application from the spread between identical
training runs.

\paragraph{Spending a budget, and training in low precision.} The data-selection literature studies
the trade of Section~\ref{sec:frontier} in another currency, holding precision fixed and varying
which examples are kept: dataset distillation \citep{wang2018dataset}, data pruning
\citep{sorscher2022beyond} and scaling laws \citep{kaplan2020scaling}. A separate literature
reduces numerical precision inside the optimiser \citep{gupta2015deep}. There precision is lost
during optimisation, and averaging can partly undo the damage. Here information is destroyed in the
stored data before optimisation starts. The two are complementary, and we make no claim about the
interaction.

\section{Limitations}
\label{sec:limitations}
\label{sec:refuted}
\label{sec:codec}

Six PDE families, two of which share initial conditions and so probe the equation only, and two of
which are one equation at distant parameters. Two further families could not be measured at all:
every run carried a gate fixed before it, and an arm whose cost at its lowest trained rate was
within $5\%$ of $1$ was declared uninformative rather than counted as a success
(Appendix~\ref{app:excluded}). Three architectures, two of them global. Two to five seeds. A single
output horizon. The probe's calibration belongs to the codec, the families and the architecture it
was measured on, and the cells carrying those figures have two seeds and intervals wide enough to
contain one another.

\paragraph{No mechanism.} The measurements are consistent and reproducible, which invites an
explanation. We tested seven and refuted all seven, each with the measurement that refutes it
(Appendix~\ref{app:refuted}). \emph{Errors adding in quadrature}, the natural derivation of
Equation~\eqref{eq:pred}, needs the compressed model's residual to match $\varepsilon_0$, and that
fails by up to $12.9\times$. \emph{Support dimension}, \emph{information spreading},
\emph{operator-aware allocation} and \emph{a heavy-tailed codebook} each failed a condition fixed
before it was computed. \emph{Codec error outside the band the operator reads} came closest: it
passed its first pre-registered condition across three codecs at $\rho = +0.826$, then reversed
sign to $\mathbf{-0.667}$ \emph{within} a codec. The seventh denies the phenomenon instead of
explaining it: field error's inversions might be a noisy measurement of a quantity that really does
order cost. An errors-in-variables model in the rank domain rules that out by a factor of fifteen.
The across-family inversions require $\hat\sigma = 2.38$ in log units, while zero inversions in
sixteen within-family pairs give $\sigma_{95} = 0.16$. Of the 36 inverted pairs, twenty-one are
separated by more than $3\times$ in field error and one by $17.8\times$, and no noise level flips
gaps of that size. Fitting the probe instead is the contrast: its 12 inversions require
$\hat\sigma = 0.40$, their median separation is $1.08\times$, and the second condition passes at a
median separation of $1.15$ against $1.80$ log units. \textbf{Field error fails on pairs it should
find easy, and the probe fails only on near-ties.}

\paragraph{A defect in our own quantiser.} Our first Lloyd--Max table had not converged at high bit
depth. Under it, advection's cost curve flattened at $2.85\times$ from half a bit per value onward,
which we first read as advection training data being incompressible. It is not: a converged
companded codebook, with the per-coefficient cap raised from $12$ to $14$ bits, brings the measured
cost at $8$ bits per value down to $\mathbf{1.27\times}$. The ordering across families survives,
since advection still fails the $10\%$ threshold at every rate tested. Every trained point in this
paper uses the corrected codebook (Appendix~\ref{app:protocol}).

\paragraph{Input-only compression.} We compress inputs and train against clean targets. A factor of
$263\times$ on the input is about $2\times$ on a stored input--target pair, and close to nothing on
a PDEBench archive, which ships two hundred timesteps of which we touch one. The ceiling is
structural: a corrupted target changes the training objective rather than the forward map, so no
forward-pass measurement can see target compression at all. Read our rates as how coarsely an input
field can be stored before the trained model suffers, and not as an archive saving.

\paragraph{Grid restriction and the fifth family.} Every two-dimensional number here comes from
Darcy restricted by keeping one point in four, which aliases. We therefore repeated the family
under area averaging, which does not, against three conditions fixed in advance. All three were
met, with downstream costs agreeing to within $\mathbf{1.08\times}$ (Appendix~\ref{app:grid}).
Shallow water trains on a tenth of the data of the others, and its nominal rates understate the
stored rate by up to $3.05\times$. Either could account on its own for the calibration move that
arrived with that family, so we report the move and do not attribute it.

\section{Conclusion}
\label{sec:conclusion}

What reaches a trained operator is the input perturbation after the operator has acted on it, and
that factor varies by more than two orders of magnitude across PDE families. Measuring it takes one
model trained at full precision and forward passes through it. The resulting probe inverts $12$ of
$104$ across-family cost comparisons against field error's $36$ under mean squared error. Under the
relative-$L^2$ objective of the reference recipe it inverts $11$ against $20$, losing no pair of
families under either. Its magnitude does not transfer across codecs or architectures, so we offer
it as a way to rank datasets and rates, not as a way to read a cost off. For a curator the
consequences are three. Set tolerances per dataset rather than per archive. Measure transmission
before choosing one. And check whether the budget is better spent on precision or on samples.

\section{Reproducibility statement}
\label{sec:repro}

Every run script, the raw result files it wrote and the figure scripts are released. Each script
carries the hypothesis it tests and the condition that would refute it. \textbf{Thirty-eight}
acceptance thresholds were written into those scripts before the runs that scored them.
\textbf{Fifteen were missed}, twenty were met, and three are void because they could not have
failed. Appendix~\ref{app:bars} lists all 38 bars, the 15 that were missed included, with the value
each would have had to reach to fail. A bar that could not have failed is not evidence. Every field
used here comes from the public PDEBench release \citep{takamoto2022pdebench}, deposited at
\url{https://doi.org/10.18419/darus-2986} under CC~BY~4.0. The checker travels with the scripts,
and compares every number in the paper's tables against the file it was computed from.

\impact{This work concerns the storage format of scientific simulation benchmarks, and we see no
significant risk of harm. The plausible positive effect is cheaper storage and transfer of
public benchmarks, together with a caution against setting compression tolerances from field
statistics alone, which our measurements indicate can understate the cost to a trained model by
a large factor.}

\acks{This work received no funding. The author declares no competing interests.}

\vskip 0.2in
\bibliography{refs}

\appendix

\section{Background, and the derivations the body omits}
\label{app:background}

A compressor here must preserve a functional of the stored field, namely the accuracy of an
operator fitted to it. Coding an observation to preserve a downstream functional is the indirect,
or remote, source coding problem \citep{wolf1970transmission}: the encoder sees $A$ and is judged
on $F(A)$, so the operative rate--distortion function is
\begin{equation}
R_F(\mathcal{D}) \;=\; \min_{\,p(\hat A \mid A)\;:\;\mathbb{E}\,d\bigl(F(A),\,F(\hat
A)\bigr)\,\le\,\mathcal{D}} I(A; \hat A) , \label{eq:remote} \end{equation}
which collapses to the ordinary $R(\mathcal{D})$ exactly when $F$ preserves the metric. Advection
is the instructive exception, a shift and therefore an isometry, which preserves the size of any
perturbation, and the only family in our set where the two distortions coincide.

The field-side answer itself is classical, and we restate it only to establish that it describes
our data well. A dataset of input-output pairs with inputs drawn from a measure $\pi$ can be
treated as a sample of the pushforward measure $F_\#\pi$. Writing $N(t) = \#\{j : v_j > t\}$ for
the counting function of the eigenvalues of $\mathrm{Cov}(F_\#\pi)$, reverse water-filling at
level $\theta$, which spends bits only on coefficients whose variance exceeds $\theta$ and
discards the rest, gives, exactly,
\begin{equation}
\mathcal{D}(\theta) = \int_0^\theta N(t)\,dt, \qquad
b(\theta) = \frac{1}{2\ln 2}\int_\theta^\infty \frac{N(t)}{t}\,dt, \qquad
-\frac{d\mathcal{D}}{db} = 2\ln 2 \cdot \theta ,
\label{eq:rd}
\end{equation}
so the water level is the marginal price of a bit. The PDE fixes $N$. For solution samples of
Sobolev regularity $m$ in $d$ spatial dimensions, drawn through an operator smoothing by order $s$
from an input measure $(-\Delta + \tau^2)^{-\alpha}$,
\begin{equation} v_j \;\asymp\; j^{-\gamma}, \qquad \gamma \;=\; 1 +
\frac{2m}{d}, \qquad m \;=\; \alpha - \frac{d}{2} + s . \label{eq:decay} \end{equation}
We write $m$ here because $r$ is the stored rate everywhere else. These identities are classical
\citep{cover2006elements}, and the sharp single-source quantisation rates they imply are due to
\citet{luschgy2004sharp}. We claim none of them, and report them because the paper's point depends
on the framework being \emph{accurate}: if it were a poor description of the data, the decoupling
of Section~\ref{sec:decouple} would be unremarkable. Appendix~\ref{app:spectral} collects our
checks, of which the sharpest is a scaling exponent satisfying $p^\star(\gamma-1) = 0.9986 \pm
0.0147$ across 8 values of $\alpha$, together with the three respects in which the framework is
loose, each measured rather than asserted and costing between about $1\%$ and $25\%$.

\paragraph{The allocation and what it is charged.} Writing $v_j$ for the variance of coefficient
$j$ and $d(b)$ for the unit-variance distortion of the $b$-bit codebook, the allocation is the
integer vector $b$ solving
\begin{equation} \min_{b \,\in\, \mathbb{Z}_{\ge
0}^{D}} \; \sum_{j=1}^{D} v_j\, d(b_j) \qquad \text{subject to} \qquad \sum_{j=1}^{D} b_j \;=\;
D r , \quad b_j \le b_{\max} . \label{eq:alloc} \end{equation}
Greedy assignment solves this exactly, because the gains $d(b) - d(b+1)$ are positive and strictly
decreasing, a property verified when the codebooks are built. This is Equation~\eqref{eq:rd} at
integer resolution. The side information the decoder needs is $b$ with the per-coefficient mean and
standard deviation of the training split, and no data-derived basis is used. Charged as
Section~\ref{sec:accounting} requires, those two vectors and the allocation table cost
$D\,(2b_{\mathrm{f}} + b_{\mathrm{a}})$ bits once for the whole family, so
\begin{equation}
r_{\mathrm{stored}} \;=\; r \;+\; \frac{2b_{\mathrm{f}} + b_{\mathrm{a}}}{n} , \qquad
b_{\mathrm{f}} = 32, \quad b_{\mathrm{a}} = 4 , \label{eq:side} \end{equation}
with $b_{\mathrm{f}}$ the width of a stored float and $b_{\mathrm{a}}$ the bits per coefficient of
the allocation table. The field size cancels, so the overhead is $68/n$ bits per value at any $D$:
$0.0068$ at $n = 10^4$ and ten times that at $n = 10^3$. For the 1D families that is a few per
cent of the payload. For 2D shallow water at the $0.03125$ bpv we trained there it exceeds the
payload. Every shallow-water rate quoted here is nominal, and Appendix~\ref{app:protocol} gives
the factor, which reaches $\mathbf{3.05\times}$.

\paragraph{The field error a tolerance allows.} Substituting $T = \kappa\, e$ in
Equation~\eqref{eq:rstar} gives the field error a curator may allow,
\begin{equation} e\bigl(r^\star(t)\bigr) \;=\; \sqrt{t^2 -
1}\;\frac{\varepsilon_0}{\kappa(r^\star)} , \label{eq:rstarfield} \end{equation}
which exposes two properties of the object. It is not a property of the dataset alone: a more
accurate model, meaning a smaller $\varepsilon_0$, demands better-preserved fields, and two
datasets can differ in $r^\star$ through $\kappa$, through $\varepsilon_0$, or both. It also
inherits whatever magnitude error the probe carries, which is why Section~\ref{sec:rstar} requires
a back-check against training.

\section{Spectral checks}
\label{app:spectral}

These measurements establish that the field-side framework of Equation~\eqref{eq:rd} is an accurate description of the data, and the decoupling in Section~\ref{sec:decouple} is
worth reporting for that reason. They come from two different sources, which we distinguish because they are not equally strong
evidence. The first is measurements on PDEBench. The second is measurements on datasets
generated by our own solver during an earlier phase of this work, where the input measure and
the smoothing order are known by construction, so the predictions can be checked against ground
truth instead of against a fit.

\subsection{Own-solver experiments, where the ground truth is known}

For an input measure $(-\Delta+\tau^2)^{-\alpha}$ and an operator smoothing by order $s$, the theory predicts $\gamma = 1+2m/d$ with $m = \alpha - d/2 + s$, hence a distortion exponent
$p^\star$ satisfying $p^\star(\gamma-1) = 1$.

\begin{itemize}
\item \textbf{Scaling law.} Across 8 values of $\alpha$, spanning $\gamma$ from $2.25$ to
$4.87$ on a two-dimensional elliptic problem, $p^\star(\gamma-1) = 0.9986 \pm 0.0147$. The
exponent must be measured on the number of resolved \emph{modes} rather than on bits: $b^\star/m^\star$
is still growing over any reachable window, so a bit-exponent is inflated.

\item \textbf{The $s$ axis, independently.} Shifting the output norm by one Sobolev order shifts
the measured $\gamma$ by $1.012$ and $1.016$, against a predicted shift of $1.000$. This tests
the $+s$ term separately from the $\alpha - d/2$ term.

\item \textbf{Nonlinearity and shocks.} A viscosity sweep on 1D Burgers, using the empirical
covariance of the pushforward so that no linearisation is required, gives $m$ falling monotonically toward the theoretical shock limit of $1/2$. At the sharpest
resolvable viscosity the converged estimate is $m = 0.523$ against $0.5$, an error of $4.6\%$. We report the
converged number: a first pass using a wide fit window gave
$\gamma = 2.240$ and hence $m = 0.620$, and an $n$-convergence check from $200$ to $1600$ samples showed the wide window
still drifting while a narrow window converged to $\gamma \approx 2.05$. Fit windows must be
truncated where eigenvalues reach numerical noise; not doing so made $\gamma$ non-monotone in
viscosity, which is how the problem was found.

\item \textbf{Dissipation identity.} For a semigroup with mode decay rates $\mu_j$, the
log-spectrum drifts at $2\mu_j$ and each mode leaves the sum as it passes the water level, giving
$-db/dt = (1/\ln 2)\,\mathrm{tr}(L\,|\,\text{resolved})$. The differential form holds to a ratio
of $1.0000$; integrated over a horizon $T_{\mathrm{h}}=3$ the predicted drop is $936.68$ bits against $936.65$
measured, a ratio of $0.999968$. Pure transport gives exactly zero, and advection--diffusion
matches pure diffusion, as it must since advection is unitary.

\item \textbf{Two regimes.} Below a crossover the distortion follows the spectral power law;
above it, once the water level has passed the whole spectrum, it becomes exponential in the rate.
The dimensional-regime slope is $2.000$ for $\gamma \in [1.02, 2]$ and every grid size tested,
and the crossover exponent measures $-0.514$, $-0.997$ and $-1.994$ against predictions of
$-0.5$, $-1.0$ and $-2.0$.
\end{itemize}

\subsection{PDEBench measurements}

On PDEBench 2D Darcy the pushforward exponent is stable across fit windows,
$\gamma_{\text{out}} = 2.456$, $2.483$, $2.516$, $2.537$ over four windows, a spread of $3\%$,
and converged in sample count: $2.532$, $2.495$, $2.486$, $2.483$ at
$n = 10^3, 2.5\times10^3, 5\times10^3, 10^4$. The input exponent is
$\gamma_{\text{in}} \approx 1.58$, giving a smoothing order
$s = \gamma_{\text{out}} - \gamma_{\text{in}} = 0.904$ against the value of $1$ expected for
Darcy flow. Note that PDEBench's coefficient field is considerably rougher than a smooth
Gaussian random field, $\gamma_{\text{in}} = 1.58$ against $2.0$ for a Karhunen--Lo\`eve field, which is near the
trace-class boundary at $\gamma = 1$ and is why its compression numbers are worse than those of
our own generated data.

The measured rate--distortion curve of the stored input field is

\begin{center}
\begin{tabular}{lrrrrrr}
\toprule
budget (bits/sample) & 64 & 128 & 256 & 512 & 1024 & 2048 \\
relative error (\%)  & 25.55 & 20.77 & 16.71 & 13.24 & 10.22 & 7.56 \\
\bottomrule
\end{tabular}
\end{center}

\noindent with a fitted slope of $-0.3484$ against the theoretical
$-(\gamma_{\text{in}}-1)/2 = -0.290$, a ratio of $1.20$. The law is right in form and about
$20\%$ off in exponent on real data; a synthetic control with a known spectrum gave $0.94$.

\subsection{Where the framework is loose}

Three known gaps, each with a measurement attached. Reverse water-filling on a covariance is exact only for a Gaussian source. Refitting the
quantiser to the empirical marginal of each coefficient recovers $4$ to $22\%$ of the
distortion, so the Gaussian surrogate costs that much. The direction of the error is
unfavourable for shocked fields, where heavy-tailed coefficients make the framework
\emph{under}state the bit cost by about $25\%$. A single encoder
shared across a multi-PDE benchmark, rather than one per equation, costs at most $10\%$; this was
measured by constructing families of operators with controlled non-commutativity, and the
obstruction is real and monotone in that non-commutativity but small. Finally, the framework is
undefined for $\gamma \le 1$, which is the trace-class boundary, and errors reach about $8\%$ at
$\gamma \approx 2$ before falling below $1\%$ for $\gamma \gtrsim 2.5$.

Taken together these gaps cost between about $1\%$ and $25\%$, depending on the axis.
The framework describes the stored field well. It is the transfer to the trained model, not the
framework itself, that fails.

\section{Protocol details}
\label{app:protocol}

What the decoder-side vectors cost family by family, and the remeasurement that put every
trained point on one quantiser.

\begin{table}[!ht]
\centering
\small
\begin{tabular}{lrrrr}
\toprule
family & $D$ & samples & side info & stored rate at the lowest trained rate \\
\midrule
1D families      & $1024$ & $10^4$ & $6.6$ bits & $0.0689$ against $0.0625$ nominal, $1.10\times$ \\
Darcy 2D         & $4096$ & $10^4$ & $26.2$ bits & $0.0689$ against $0.0625$, $1.10\times$ \\
shallow water 2D & $4096$ & $10^3$ & $262.1$ bits & $\mathbf{0.0953}$ against $0.03125$, $\mathbf{3.05\times}$ \\
\bottomrule
\end{tabular}
\caption{What the decoder-side mean and standard deviation vectors cost, charged as
Section~\ref{sec:accounting} requires. Shallow water pays most, having ten times fewer samples
to amortise over.}
\label{tab:sidecost}
\end{table}
\FloatBarrier
\paragraph{The single-protocol remeasurement.} Section~\ref{sec:codec} shows that our original
Lloyd--Max table was unconverged at high bit depth. An earlier version of this paper assembled
its sixteen trained points from both tables, with advection, shallow water and the far-parameter
set on the corrected codebook and Burgers, reaction--diffusion and Darcy on the original one; on
that assembly field error inverted $41$ of the $104$ across-family pairs and the probe $13$. We
remeasured the Burgers and reaction--diffusion cells under the corrected codebook, changing
nothing else, and took Darcy from the decimation arm of Appendix~\ref{app:grid}, which had
measured both of its rates on the corrected codebook. Three bars were written into the run
script before it executed (Table~\ref{tab:bars}). The separation had to survive, with field
error inverting at least twice as many pairs as the probe and the probe losing at most one
block; it did, at $36$ against $12$ with no block. The full-precision errors had to
reproduce within $10\%$; they did, within $0.34\%$. The field errors at the remeasured cells had
to reproduce within $15\%$, and this bar was missed. Three of the four cells moved by at most
$1.7\%$, but reaction--diffusion at half a bit per value fell from $1.370\%$ to $0.673\%$ field
error and from $1.13\times$ to $1.00\times$ cost, so that cell of the earlier table was an
artefact of the quantiser. The defect therefore reaches low \emph{rates}: reverse water-filling
gives the leading coefficients far more than 8 bits each even at $0.25$ bits per value, where
the corrected codebook moves advection from $3.29\times$ to $3.03\times$ and reaction--diffusion
from $1.33\times$ to $1.05\times$. Table~\ref{tab:decouple} was remeasured for that reason, and
the orderings it shows are unchanged.

\paragraph{Architectures and training.} The Fourier neural operator uses $16$ modes and width
$64$ in one dimension and $12$ modes and width $32$ in two. The DeepONet has a dense branch over
the whole input field, a trunk over coordinates and $p = 128$. The U-Net has four pooling levels
at base width $32$. Training is $150$ epochs in 1D and $100$ in 2D, Adam at $10^{-3}$ with cosine
annealing, and model selection on the validation split.

\paragraph{The bit-identical initial conditions.} The first 16 hex digits of the SHA-1 hash of
the full $10^4 \times 1024$ input array is \texttt{96ad83ca2c8c83d1} for both 1D advection and 1D
Burgers, against \texttt{74a4221c1260a397} for reaction--diffusion.

\section{Seed uncertainty}
\label{app:seeds}
 Over five seeds the full-precision errors are
$0.8548 \pm 0.0095\%$ (advection) and $0.0972 \pm 0.0006\%$ (reaction--diffusion), and the
measured costs carry relative standard deviations of $1$ to $7\%$ (Table~\ref{tab:sz3}). The
worst-case factors reported in this paper, $1.4$ to $1.6$, are therefore not seed noise.
Elsewhere we quote the full-precision error of the particular run being described instead of
this mean, so advection appears as $0.851\%$, $0.855\%$ and $0.857\%$ in different places; all three are single seeds of the same configuration and all lie inside the spread
above.

Two natural alternatives fail much worse and are worth recording. Assuming the codec error and
the model error are independent and add in quadrature, using the \emph{field} error, is accurate
to $7\%$ on two families and wrong by a factor of $16$ on reaction--diffusion, which is
precisely the family whose operator does the most projecting. Replacing the trained model with a ridge linear surrogate fails by a factor of $15$, and its
failure pattern is diagnostic: it is accurate exactly when the surrogate is itself competitive
with the target model. That held on advection, where the ridge fit reached $0.165\%$ against the
network's $0.851\%$, and failed on Burgers, where it reached $51.4\%$ against $9.30\%$.

\subsection{A second tier, and a pre-registration that failed}
\label{sec:prereg}

The probe's second error pattern runs along a different axis from the codec dependence above:
holding the codec fixed and varying the family, it \emph{under}-predicts on advection and on 2D
Darcy. Under-prediction means a model \emph{trained} on compressed data is worse than a clean
model merely \emph{evaluated} on it, which a forward pass cannot see by construction. We therefore built a second, still inexpensive tier, a $15$-epoch fine-tune of the full-precision model on the
compressed data against a fine-tuned control, and combined it with the probe by a geometric mean.
On the eight measurements that suggested the rule the combination was much tighter than either
tier, at $1.14\times$ worst case, so we froze it and pre-registered a $1.30\times$ bar on families
that took no part in choosing it. \textbf{The bar was missed, at $1.43\times$.} The in-sample
$1.14\times$ was optimism of exactly the kind the pre-registration was written to catch. The
geometric mean remains the best of the three tiers out of sample and we report $1.4$ to $1.6$
rather than $1.14$ as its accuracy. Appendix~\ref{app:tiers} gives the construction, the control
ratios and the per-tier table; nothing later in this paper depends on the second tier.

\subsection{The result is not specific to one architecture}
\label{sec:arch}

DeepONet shares no inductive bias with the Fourier neural operator. It reads the whole input
field through a dense branch and has no spectral structure anywhere, whereas band-limited codec
error is exactly the kind of perturbation a truncated Fourier transform is built to be smooth
in. On our measurements DeepONet is predicted as well as or better than the Fourier neural
operator, at ratios $1.09$, $0.87$, $1.19$ and $1.05$. This is weaker evidence than it appears. DeepONet's full-precision error on advection is $8.57\%$ against the Fourier operator's
$0.857\%$, and on shallow water $4.22\%$ against $0.41\%$, so by the argument of
Appendix~\ref{app:structure} its cells lie where compression is nearly invisible and any
predictor bounded below by one is accurate. Its shallow-water costs are $1.002$, $0.996$ and $0.998$.

The DeepONet arm establishes only that the probe is not \emph{worse} on a second architecture,
not that the result is architecture-independent. The worst cell over all eight is a Fourier
operator cell, so the calibration we quote is not set by the second architecture. Both of those architectures are global: one mixes through a truncated Fourier
transform and the other through a dense branch net. Neither tests what happens under a local
inductive bias.

\section{Conclusions we withdrew}
\label{app:corrections}

Each rule of Section~\ref{sec:accounting} is stated because we violated it first. This appendix
gives the violations in full.

\paragraph{Charge every data-derived basis to the bit budget.} An earlier version of this work
reported a large compression factor using a PCA basis whose storage was never counted. A
$4096\times4096$ single-precision basis amortised over $10^4$ samples costs $53{,}687$ bits per
sample, which is $41\%$ of the full-precision field itself, and the figure fell from $128\times$
to about $2.4\times$ once it was counted.

\paragraph{Hold the rate fixed, not the distortion.} Our first comparison across codecs matched
field error and never checked the rate. Interpolating the competitor's own rate--distortion curve
afterwards showed it had used $215$--$221$ bits per sample against our $64$. The conclusion drawn
from it was withdrawn and is re-derived at equal bits in Section~\ref{sec:ratefair}, where it
reverses. For a question about storage, rate is the axis that must be held fixed, and it should
be printed in the table of experimental arms.

\paragraph{State the objective, because it moves the baseline.} Every model here was trained by
minimising mean squared error on globally normalised fields while being evaluated in relative
$L^2$, and an earlier version of this paper did not say so. The reference Fourier neural
operator recipe trains in relative $L^2$ on the physical field. Retraining Burgers that way
brings its full-precision error from $9.275\%$ to $\mathbf{3.577\%}$, a factor of $2.6$, and its
cost at $0.125$ bits per value from $1.07\times$ to $1.62\times$. This matters beyond the
number: the explanation Section~\ref{sec:decouple} gives for Burgers' cheapness rests on that
$9.3\%$ baseline, and is therefore substantially an artefact of an objective we had not stated.
The ordering of the family-level median transmission factors is the same under either objective, but a baseline we used as an explanation is under two fifths as large under the
standard one. Appendix~\ref{app:objective} retrains all sixteen cells of
Table~\ref{tab:decouple8} under relative $L^2$; the rest of the paper is reported under the
objective it was run with, and we flag every place the Burgers baseline is load bearing.

\paragraph{Check that the quantiser converged.} The Lloyd--Max table used in our earlier runs
initialises levels on $[-2.5, 2.5]$ and runs a fixed number of iterations, which does not spread
$4096$ levels out to $\pm 4\sigma$. Its distortion at $12$ bits is $7.1\times10^{-6}$ against
$1.6\times10^{-7}$ for a companded codebook on the same density, a factor of $44$. Section~\ref{sec:codec} shows this alone changed a headline conclusion.

\section{The trained measurements, by block and by rate}
\label{app:blocks}

\begin{figure}[!ht]
\centering
\includegraphics{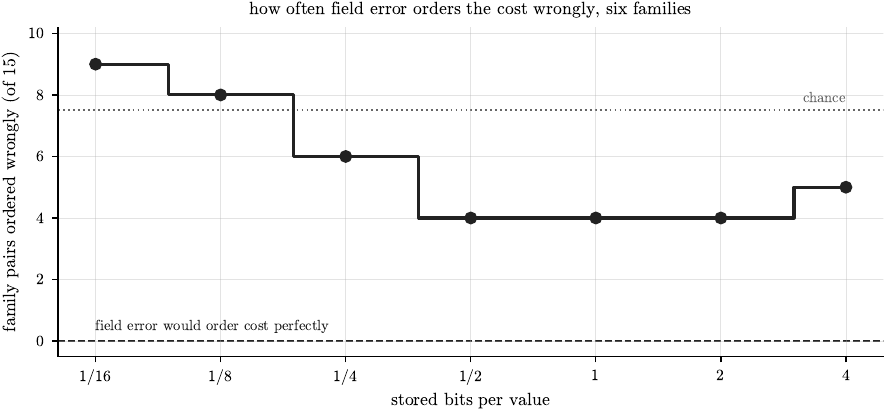}
\caption{At each of the seven rates all six families share, how many of the 15 family pairs
field error orders the wrong way, against the $7.5$ that chance would give. Costs are
probe-curve values.}
\label{fig:decouple}
\end{figure}

Table~\ref{tab:decouple8} is every point in this work at which a model was trained on compressed
data. Table~\ref{tab:blocks} breaks the 104 across-family pairs of Section~\ref{sec:decouple} out by
family pair, and Figure~\ref{fig:decouple} sweeps the same count over rates the probe reaches
and training cannot. It inverts $9$ at a sixteenth of a bit per value and at least $4$ at every
rate, so the failure is not an artefact of the rates we happened to train at.

\begin{table}[!ht]
\centering
\footnotesize
\begin{tabular}{llrr}
\toprule
family & rate (bpv) & field error & measured cost \\
\midrule
advection           & $4.0$     & $0.379\%$  & $1.19\times$ \\
reaction--diffusion & $0.5$     & $0.673\%$  & $1.00\times$ \\
advection           & $1.0$     & $0.762\%$  & $1.52\times$ \\
reaction--diffusion & $0.125$   & $4.278\%$  & $2.57\times$ \\
Burgers             & $0.125$   & $5.746\%$  & $1.07\times$ \\
Darcy 2D            & $0.25$    & $13.540\%$ & $\mathbf{1.27\times}$ \\
Darcy 2D            & $0.0625$  & $19.565\%$ & $1.73\times$ \\
Burgers             & $0.03125$ & $35.082\%$ & $2.28\times$ \\
\midrule
shallow water 2D    & $0.25$    & $3.900\%$  & $1.65\times$ \\
shallow water 2D    & $0.125$   & $4.478\%$  & $2.05\times$ \\
shallow water 2D    & $0.0625$  & $5.055\%$  & $2.90\times$ \\
shallow water 2D    & $0.03125$ & $5.790\%$  & $3.06\times$ \\
\midrule
reaction--diffusion, far $\nu,\rho$ & $0.5$     & $0.687\%$  & $1.15\times$ \\
reaction--diffusion, far $\nu,\rho$ & $0.25$    & $1.818\%$  & $1.80\times$ \\
reaction--diffusion, far $\nu,\rho$ & $0.125$   & $4.303\%$  & $3.05\times$ \\
reaction--diffusion, far $\nu,\rho$ & $0.0625$  & $10.202\%$ & $6.92\times$ \\
\bottomrule
\end{tabular}
\caption{Every trained measurement: six families, corrected codebook, two seeds per cell. The
last block is reaction--diffusion at $\nu = 0.5$, $\rho = 10$, a second dataset of an equation
already present.}
\label{tab:decouple8}
\end{table}

\begin{table}[!ht]
\centering
\small
\begin{tabular}{lrrr}
\toprule
family pair & point pairs & field error inverts & the probe inverts \\
\midrule
advection $\times$ shallow water & $8$ & $\mathbf{0}$ & $0$ \\
Burgers $\times$ Darcy 2D & $4$ & $\mathbf{0}$ & $1$ \\
reaction--diffusion $\times$ far $\nu,\rho$ & $8$ & $\mathbf{0}$ & $1$ \\
advection $\times$ far $\nu,\rho$ & $8$ & $1$ & $1$ \\
reaction--diffusion $\times$ shallow water & $8$ & $1$ & $1$ \\
advection $\times$ reaction--diffusion & $4$ & $1$ & $0$ \\
advection $\times$ Darcy 2D & $4$ & $1$ & $2$ \\
advection $\times$ Burgers & $4$ & $2$ & $0$ \\
Burgers $\times$ reaction--diffusion & $4$ & $2$ & $0$ \\
reaction--diffusion $\times$ Darcy 2D & $4$ & $2$ & $0$ \\
shallow water $\times$ far $\nu,\rho$ & $16$ & $3$ & $3$ \\
Burgers $\times$ shallow water & $8$ & $5$ & $1$ \\
Burgers $\times$ far $\nu,\rho$ & $8$ & $5$ & $1$ \\
Darcy 2D $\times$ far $\nu,\rho$ & $8$ & $\mathbf{6}$ & $0$ \\
Darcy 2D $\times$ shallow water & $8$ & $\mathbf{7}$ & $1$ \\
\bottomrule
\end{tabular}
\caption{The 15 family pairs, sorted by how often field error inverts them. Counting one vote
per block rather than one per point pair, field error loses 4 blocks and the probe loses none.}
\label{tab:blocks}
\end{table}
\FloatBarrier

\begin{table}[!ht]
\centering
\small
\begin{tabular}{lrrrr}
\toprule
predictor & within a family & across families & $\rho$ & worst magnitude \\
\midrule
field reconstruction error & $0$ of $16$ & $36$ of $104$ & $+0.48$ & --- \\
$\sqrt{1+(\text{field error}/\varepsilon_0)^2}$ & $0$ of $16$ & $29$ of $104$ & $+0.66$ & $71.2\times$ \\
$\varepsilon_0$ alone, no codec information & $0$ of $16$ & $38$ of $104$ & $+0.31$ & --- \\
transmitted error $T$ alone & $0$ of $16$ & $53$ of $104$ & $+0.13$ & --- \\
the probe, $\sqrt{1+(T/\varepsilon_0)^2}$ & $0$ of $16$ & $\mathbf{12}$ of $104$ & $\mathbf{+0.93}$ & $\mathbf{2.11\times}$ \\
\bottomrule
\end{tabular}
\caption{Taking the probe apart on the 16 trained measurements. Every predictor scores $0$ of
$16$ within a family, so only the across-family column discriminates.}
\label{tab:ablation}
\end{table}

\begin{table}[!ht]
\centering
\small
\begin{tabular}{lrrrr}
\toprule
families & pairs & field error inverts & the probe inverts & blocks lost \\
\midrule
all six & $104$ & $36$ ($35\%$) & $\mathbf{12}$ ($12\%$) & $4$ vs $\mathbf{0}$ of $15$ \\
without advection & $76$ & $31$ ($41\%$) & $9$ ($12\%$) & $4$ vs $0$ of $10$ \\
without Burgers & $76$ & $22$ ($29\%$) & $9$ ($12\%$) & $2$ vs $0$ of $10$ \\
without reaction--diffusion & $76$ & $30$ ($39\%$) & $10$ ($13\%$) & $4$ vs $0$ of $10$ \\
without Darcy 2D & $76$ & $20$ ($26\%$) & $8$ ($11\%$) & $2$ vs $0$ of $10$ \\
without shallow water & $56$ & $20$ ($36\%$) & $6$ ($11\%$) & $2$ vs $0$ of $10$ \\
without far $\nu,\rho$ & $56$ & $21$ ($38\%$) & $6$ ($11\%$) & $2$ vs $0$ of $10$ \\
without advection and Burgers & $52$ & $19$ ($37\%$) & $6$ ($12\%$) & $2$ vs $0$ of $6$ \\
without shallow water and far $\nu,\rho$ & $24$ & $8$ ($33\%$) & $3$ ($13\%$) & $0$ vs $0$ of $6$ \\
\bottomrule
\end{tabular}
\caption{The across-family comparison of Figure~\ref{fig:headtohead} with families removed: each
in turn, the two that share initial conditions, and the two added last. Blocks lost counts the
family pairs on which a predictor inverts more than half of the point pairs, field error first.}
\label{tab:robust}
\end{table}
\section{The training objective}
\label{app:objective}

Section~\ref{sec:probe} summarises this appendix. Every other model in the paper minimises mean
squared error on globally normalised fields. The sixteen cells of Table~\ref{tab:decouple8} were
retrained twice more, with data, splits, codebook and cap, architecture, epochs, optimiser,
schedule, seeds, validation-based model selection, probe and rates all unchanged.

The first retraining computed relative $L^2$ on the normalised targets,
$\lVert P - U_n \rVert / \lVert U_n \rVert$ with $U_n = (U - m_U)/s_U$, where $m_U$ and $s_U$ are
the global mean and standard deviation of the training targets; an earlier Burgers run had used the
same form. Three bars were fixed before it ran: that field error invert at least twice as many
across-family pairs as the probe, with the probe losing at most one block; that the per-family
median transmission factor keep its order, at a rank correlation of at least $0.9$; and that the
field errors and the Burgers baseline reproduce. The first was missed, at $23$ against $19$, and
the other two were met. Only after scoring did we compare the loss with the reference
implementation, which decodes the prediction and computes relative $L^2$ on the physical field.
The two forms coincide for the zero-mean fields of advection and Burgers. For the other four
families, subtracting a large global mean leaves some samples with small normalised norms and large
weights in the loss, and the two reaction--diffusion families trained unstably under it, their
seeds' full-precision errors $46\%$ and $49\%$ apart.

The second retraining computed $\lVert P\,s_U + m_U - U \rVert / \lVert U \rVert$, under the same
three bars and a fourth written for the diagnosis: that both reaction--diffusion families train
stably, with the larger seed's full-precision error within $15\%$ of the smaller. The diagnosis
held, at $1.1\%$ and $2.1\%$, and so did the ordering and consistency bars; the separation
was missed again, at $20$ against $11$. Leaving out any one family under the physical form leaves
field error inverting $13$ to $24\%$ of the remaining across-family pairs and the probe
$5$ to $14\%$, and the probe loses no block in any subset; with shallow water and the
far-parameter set both removed it inverts $4$ of the $24$ remaining pairs against field
error's $3$. Table~\ref{tab:objective} summarises the three runs, Table~\ref{tab:objeps} gives the
full-precision errors and Table~\ref{tab:objcells} every cell.

\begin{table}[!ht]
\centering
\small
\begin{tabular}{lrrrrr}
\toprule
objective & field error & the probe & blocks lost & $\kappa$ order & worst $\rho$ \\
\midrule
mean squared error & $36$ of $104$ & $12$ of $104$ & $4$ vs $0$ & --- & $2.11\times$ \\
relative $L^2$, normalised targets & $23$ of $104$ & $19$ of $104$ & $1$ vs $0$ & $1.00$ & $2.88\times$ \\
relative $L^2$, physical field & $20$ of $104$ & $\mathbf{11}$ of $104$ & $1$ vs $0$ & $1.00$ & $1.72\times$ \\
\bottomrule
\end{tabular}
\caption{The sixteen trained cells under three training objectives, everything else fixed.
$\kappa$ order is the rank correlation of each family's median transmission factor with its
value under mean squared error, and worst $\rho$ is Equation~\eqref{eq:rho} over the sixteen
cells.}
\label{tab:objective}
\end{table}

\begin{table}[!ht]
\centering
\footnotesize
\begin{tabular}{lrrr}
\toprule
family & mean squared error & relative $L^2$, normalised & relative $L^2$, physical \\
\midrule
advection & $0.8570\%$ & $0.7267, 0.7256\%$ & $0.7323, 0.7193\%$ \\
Burgers & $9.2747\%$ & $3.4380, 3.4438\%$ & $3.4880, 3.6654\%$ \\
reaction--diffusion & $0.0972\%$ & $\mathbf{0.0847, 0.0581\%}$ & $0.0846, 0.0856\%$ \\
Darcy 2D & $4.2154\%$ & $2.3740, 2.3652\%$ & $2.4750, 2.4803\%$ \\
shallow water 2D & $0.4099\%$ & $0.2903, 0.3017\%$ & $0.4027, 0.3852\%$ \\
reaction--diffusion, far $\nu,\rho$ & $0.0207\%$ & $\mathbf{0.0452, 0.0304\%}$ & $0.0260, 0.0265\%$ \\
\bottomrule
\end{tabular}
\caption{Full-precision test error under the three objectives: the mean of two seeds for mean
squared error, each seed for relative $L^2$. Bold marks the two families the normalised form
left far apart.}
\label{tab:objeps}
\end{table}

\begin{table}[!ht]
\centering
\footnotesize
\begin{tabular}{llrrrrr}
\toprule
family & rate & field error & cost, MSE & normalised & physical & probe \\
\midrule
advection & $1.0$ & $0.762\%$ & $1.52\times$ & $1.12\times$ & $1.11\times$ & $1.42\times$ \\
advection & $4.0$ & $0.379\%$ & $1.19\times$ & $1.03\times$ & $1.03\times$ & $1.16\times$ \\
Burgers & $0.03125$ & $35.082\%$ & $2.28\times$ & $6.41\times$ & $5.93\times$ & $9.96\times$ \\
Burgers & $0.125$ & $5.746\%$ & $1.07\times$ & $1.64\times$ & $1.62\times$ & $1.83\times$ \\
reaction--diffusion & $0.125$ & $4.278\%$ & $2.57\times$ & $1.75\times$ & $2.49\times$ & $4.30\times$ \\
reaction--diffusion & $0.5$ & $0.673\%$ & $1.00\times$ & $1.14\times$ & $0.98\times$ & $1.03\times$ \\
Darcy 2D & $0.0625$ & $19.565\%$ & $1.73\times$ & $2.61\times$ & $2.69\times$ & $1.77\times$ \\
Darcy 2D & $0.25$ & $13.540\%$ & $1.27\times$ & $1.71\times$ & $1.75\times$ & $1.29\times$ \\
shallow water 2D & $0.03125$ & $5.790\%$ & $3.06\times$ & $4.33\times$ & $3.25\times$ & $3.81\times$ \\
shallow water 2D & $0.0625$ & $5.055\%$ & $2.90\times$ & $3.71\times$ & $2.76\times$ & $3.50\times$ \\
shallow water 2D & $0.125$ & $4.478\%$ & $2.05\times$ & $3.33\times$ & $2.15\times$ & $3.33\times$ \\
shallow water 2D & $0.25$ & $3.900\%$ & $1.65\times$ & $2.69\times$ & $1.82\times$ & $2.47\times$ \\
reaction--diffusion, far $\nu,\rho$ & $0.0625$ & $10.202\%$ & $6.92\times$ & $3.37\times$ & $11.53\times$ & $11.56\times$ \\
reaction--diffusion, far $\nu,\rho$ & $0.125$ & $4.303\%$ & $3.05\times$ & $1.16\times$ & $1.85\times$ & $3.03\times$ \\
reaction--diffusion, far $\nu,\rho$ & $0.25$ & $1.818\%$ & $1.80\times$ & $1.26\times$ & $1.15\times$ & $1.19\times$ \\
reaction--diffusion, far $\nu,\rho$ & $0.5$ & $0.687\%$ & $1.15\times$ & $1.14\times$ & $1.01\times$ & $1.02\times$ \\
\bottomrule
\end{tabular}
\caption{Every cell under the three objectives. Field error does not depend on the objective,
and each cost is relative to the full-precision model of its own objective.}
\label{tab:objcells}
\end{table}
\FloatBarrier
\section{Structural quantities}
\label{app:structure}

\begin{figure}[!ht]
\centering
\includegraphics{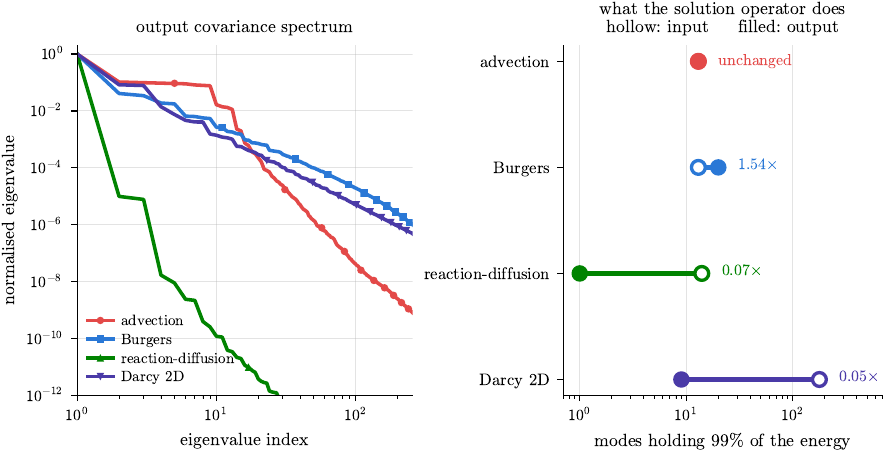}
\caption{Left: eigenvalue spectrum of the output covariance, $4000$ samples per family,
normalised by the leading eigenvalue. Right: how many modes hold $99\%$ of the energy, before
and after the solution operator.}
\label{fig:spectra}
\end{figure}

Section~\ref{sec:rstar} states that $r^\star$ does not track the operator's contraction or the
spectral decay exponents. Table~\ref{tab:structure} is the evidence. $\gamma$ is fitted on the sample covariance over a single window truncated at numerical noise,
so Darcy's $\gamma_{\text{in}}$ reads $1.562$ here against the $1.58$ of
Appendix~\ref{app:spectral}, where the spread across four windows is reported and is about
$3\%$. The effective rank is the participation ratio of the eigenvalue spectrum, and the
contraction ratio is the output effective rank divided by the input one. This is a different measure from the mode count of
Figure~\ref{fig:spectra}, and for Burgers the two disagree in direction, as
Section~\ref{sec:decouple} notes: its output is dominated by a single mode, which the
participation ratio sees, behind a long tail, which the $99\%$ threshold sees. Ordering by either
measure is non-monotone in $r^\star$.

\begin{table}[!ht]
\centering
\begin{tabular}{lrrrrrrr}
\toprule
family & dim & $\gamma_{\text{in}}$ & $\gamma_{\text{out}}$ & eff.\ rank in & out &
contraction & $r^\star(1.10)$ \\
\midrule
advection & 1 & $5.165$ & $5.442$ & $2.9$ & $2.9$ & $0.9999$ & $>\!8$ \\
Burgers & 1 & $5.165$ & $2.232$ & $2.9$ & $1.3$ & $0.4577$ & $0.1217$ \\
reaction--diffusion & 1 & $5.124$ & $5.037$ & $10.8$ & $1.0$ & $0.0923$ & $0.4080$ \\
Darcy & 2 & $1.562$ & $2.473$ & $8.3$ & $1.4$ & $0.1745$ & $0.1647$ \\
\bottomrule
\end{tabular}
\caption{Structural quantities against the candidate rate. Ordering by contraction is not
monotone in $r^\star$, and with four families we fit nothing to this table.}
\label{tab:structure}
\end{table}
\FloatBarrier

The reason no law should be fitted is that $r^\star$ is a level crossing, not an independent
property: at $r^\star(1.10)$ the ratio $T/\varepsilon_0$ is $\sqrt{1.1^2-1} = 0.458$ by
construction. Table~\ref{tab:crossing} gives $r^\star$ at three tolerances.

\begin{table}[!ht]
\centering
\begin{tabular}{lrrrr}
\toprule
family & fp32 error (\%) & $r^\star(1.05)$ & $r^\star(1.10)$ & $r^\star(1.25)$ \\
\midrule
advection & $0.8570$ & $>\!8$ & $>\!8$ & $1.2800$ \\
Burgers & $9.2747$ & $0.1604$ & $0.1217$ & $0.1048$ \\
reaction--diffusion & $0.0972$ & $0.4659$ & $0.4080$ & $0.2834$ \\
Darcy & $4.2154$ & $0.3954$ & $0.1647$ & $0.0578$ \\
\bottomrule
\end{tabular}
\caption{$r^\star$ in bits per value at three tolerances, on the corrected probe curves. Advection
reaches only the loosest, because its curve flattens at $1.11\times$.}
\label{tab:crossing}
\end{table}
\FloatBarrier

Because $r^\star$ is where $T(r)$ crosses a fixed multiple of $\varepsilon_0$, a family whose
full-precision model is already inaccurate tolerates a coarser codec in relative terms.
Excluding advection, whose curve is pinned by the per-coefficient cap of
Appendix~\ref{app:ratefair}, $r^\star$ falls monotonically as $\varepsilon_0$ rises: $0.097\%
\to 0.4080$, $4.215\% \to 0.1647$, $9.275\% \to 0.1217$. This is close to tautological given Equation~\eqref{eq:rstarfield}, but it is the reason
$r^\star$ cannot be read as a property of a dataset in isolation, and it is why we report $r^\star$ alongside $\varepsilon_0$ throughout.

\section{How the probe's calibration moved}
\label{app:probe}

Section~\ref{sec:probe} reports that the probe's magnitude error grew with every family,
codec and architecture added, and stops quoting a single figure because of it. This appendix is the history behind that decision, arm by arm, with the bar each arm was
measured against.

\begin{figure}[!ht]
\centering
\includegraphics{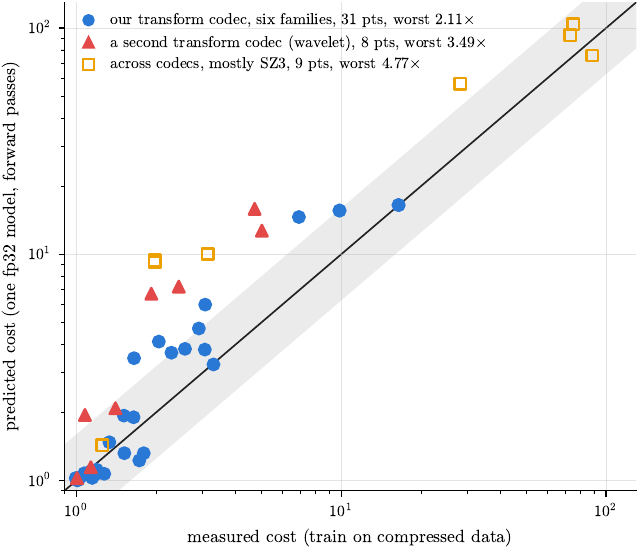}
\caption{Predicted against measured cost at all 48 points, grouped by codec: circles our
transform codec, triangles the Daubechies-4 wavelet, open squares the cross-codec arms. The
shaded band is the $1.6\times$ worst case of the first four families.}
\label{fig:probe}
\end{figure}

That $1.61\times$ does not survive a fifth family. On 2D shallow water, under the same codec and
the same architecture, the probe over-predicts at all four rates we trained, by $1.62$, $1.95$,
$2.01$ and $2.10\times$.

A sixth family missed the bar, at $2.11\times$ against a pre-registered $2.10\times$. The bar
had been set at the five-family figure precisely to test whether it was still drifting. The margin is half a per cent, well inside the $1$ to $7\%$ seed spread of Appendix~\ref{app:seeds},
so we do not read it as growth. Neither do we read it as a plateau.

That $2.11\times$ comes from a two-seed cell whose own $95\%$ interval is $[1.60, 3.11]$, and the $1.62\times$ cell behind the
five-family figure has an interval of $[1.02, 3.98]$. What is established is that the point estimate went $1.61$ to $2.10$ and then did not move
\emph{with a further family}; it did move with a further architecture, to $2.85\times$, which Appendix~\ref{sec:arch} reports. What is not established is where the worst case actually lies, and two seeds cannot establish
it. The fifth family also differs from the
others in two ways we only found later: it has a tenth of the training data
(Section~\ref{sec:setup}) and its nominal rates understate the stored rate by up to $3.05\times$
(Table~\ref{tab:sidecost}). Either could account for the move from $1.61$ to $2.10$ on its own,
so we do not attribute it to the arrival of a fifth family.

The same $1.61\times$ appears again in Table~\ref{tab:tiers} from a disjoint set of eight
out-of-sample points. This is not one number reused: the two sets share no measurement, and the
worst cell in each is Burgers at $0.03125$ bits per value, at $1.6096$ and $1.6079$ in two
separately executed runs at the same fixed seeds. The worst case of the method is therefore
reproducible across re-executions,
and it is always the same configuration.

\begin{figure}[!ht]
\centering
\includegraphics{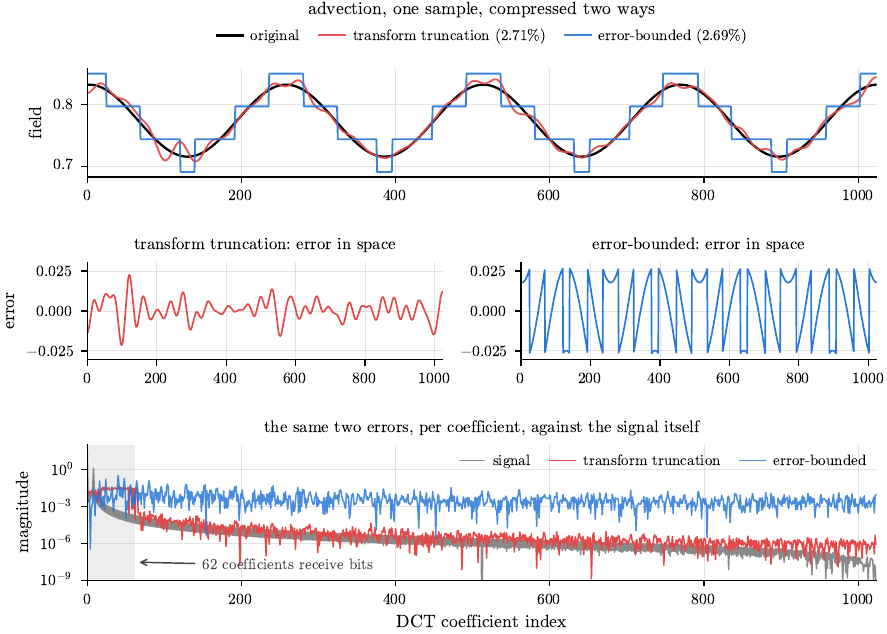}
\caption{One advection sample at $0.25$ bits per value under both codecs, the error-bounded
tolerance tuned to match field error. Top: reconstructions. Middle: errors in space, on a shared
scale. Bottom: errors per DCT coefficient.}
\label{fig:errstruct}
\end{figure}

\paragraph{Error structure.} Figure~\ref{fig:errstruct} shows one advection sample compressed
both ways at matched rate and matched field error, so the only quantity separating them is how
each spends its error budget. Transform truncation gives bits to $62$ coefficients and leaves
the remaining $962$ at their mean, so its error follows the signal. Error-bounded compression
holds a fixed tolerance at every grid point, which lays a roughly uniform residual across all
$1024$ coefficients: beyond coefficient $256$, where the signal has decayed to
$1.4\times10^{-6}$, the transform error is $3.3\times10^{-6}$ and the error-bounded error is
$3.9\times10^{-3}$. Matched field error therefore hands the network perturbations of a different
shape, and a calibration measured inside one codec has no reason to hold inside the other.
Whether error shape predicts the size of the over-prediction was not tested;
Section~\ref{sec:refuted} reports the structural explanations that were.

\paragraph{Codec dependence.} We
pre-registered a second test: sweep the tolerance of SZ3 rather than the rate of the transform
codec, and require the probe to come in under $2.0\times$, allowing slack against the
$1.61\times$ measured for the transform codec. It reached $4.65\times$ and the bar was missed.
All four points over-predict, in the same direction as the cross-codec failure
(Table~\ref{tab:sz3}).

We read that as a statement about codec \emph{classes}: the probe over-predicts for per-sample
bounded-error codecs and is accurate for transform truncation. That reading was wrong, and a
third codec showed it. We built a wavelet codec, changing the transform from the discrete cosine
to a multi-level Daubechies-4 and holding the codebook, the allocation, the architecture and the
schedule fixed, and pre-registered two bars taken from the DCT codec's own figures with no
slack: worst case at or under $1.61\times$, median at or under $1.25\times$. \textbf{Both were
missed}, at $3.49\times$ and $2.14\times$, and all 8 points over-predict.
The lower block of Table~\ref{tab:sz3} gives them. A wavelet is a transform codec by
any definition, so what the
probe is accurate for is not a class of codec but the particular one we built. The class reading is therefore refuted rather than narrowed, since it is the reading
Figure~\ref{fig:errstruct} otherwise invites.

\begin{table}[!ht]
\centering
\footnotesize
\begin{tabular}{llrrrrr}
\toprule
family & arm & rate (bpv) & field error & probe & measured & ratio \\
\midrule
\multicolumn{7}{l}{\emph{our transform codec, for reference}} \\
advection & transform & $0.2500$ & $2.743\%$ & $2.98\times$ & $3.03 \pm 0.21\times$ & $0.99$ \\
reaction--diffusion & transform & $0.1250$ & $4.278\%$ & $3.81\times$ & $2.58 \pm 0.10\times$ & $1.48$ \\
\midrule
\multicolumn{7}{l}{\emph{SZ3, an error-bounded compressor}} \\
advection & $35\%$ target & $0.1009$ & $34.49\%$ & $29.49\times$ & $12.39 \pm 0.86\times$ & $2.38$ \\
advection & $12\%$ target & $0.2468$ & $11.87\%$ & $7.17\times$ & $1.54 \pm 0.02\times$ & $\mathbf{4.65}$ \\
reaction--diffusion & $35\%$ target & $0.0634$ & $35.03\%$ & $85.14\times$ & $43.90 \pm 1.80\times$ & $1.94$ \\
reaction--diffusion & $12\%$ target & $0.1823$ & $12.01\%$ & $11.56\times$ & $5.48 \pm 0.19\times$ & $2.11$ \\
\midrule
\multicolumn{7}{l}{\emph{a Daubechies-4 wavelet, a second transform codec}} \\
advection & $0.0625$ bpv & $0.0625$ & $14.258\%$ & $15.87\times$ & $4.71\times$ & $3.37$ \\
advection & $0.125$ bpv & $0.1250$ & $7.088\%$ & $7.19\times$ & $2.44\times$ & $2.95$ \\
advection & $0.25$ bpv & $0.2500$ & $2.606\%$ & $2.08\times$ & $1.40\times$ & $1.49$ \\
advection & $0.5$ bpv & $0.5000$ & $0.896\%$ & $1.14\times$ & $1.13\times$ & $1.01$ \\
reaction--diffusion & $0.0625$ bpv & $0.0625$ & $11.230\%$ & $12.71\times$ & $5.01\times$ & $2.54$ \\
reaction--diffusion & $0.125$ bpv & $0.1250$ & $4.525\%$ & $6.69\times$ & $1.92\times$ & $\mathbf{3.49}$ \\
reaction--diffusion & $0.25$ bpv & $0.2500$ & $1.902\%$ & $1.95\times$ & $1.08\times$ & $1.81$ \\
reaction--diffusion & $0.5$ bpv & $0.5000$ & $0.613\%$ & $1.02\times$ & $1.00\times$ & $1.02$ \\
\bottomrule
\end{tabular}
\caption{The probe inside two codecs that are not ours, against the two control rows it was
calibrated on. Five seeds for the transform arms, three for SZ3 and two for the wavelet.}
\label{tab:sz3}
\end{table}
\FloatBarrier

\paragraph{Cross-codec behaviour.} Every one of the 12 non-control rows of Table~\ref{tab:sz3}
over-predicts. Figure~\ref{fig:probe} leaves out the two arms whose feasibility gate declared
them uninformative, which would otherwise pile up at the origin and make the probe look exact on
data that cannot test it. The nine cross-codec points carry a median magnitude error of $2.01\times$, too
large to offer the probe as a cross-codec predictor of cost. Their ordering is a separate matter and it survives: four of the
thirty-six pairs are ranked wrongly, at a rank correlation of $0.92$ with $p = 0.0005$. A curator therefore cannot read an absolute cost off the probe when the codec changes, and can
still use it to choose between two codecs at a fixed rate. We report the calibration failure as the limitation because it is what matters when the tool is
used as intended, to pick a rate.

\paragraph{A third architecture.} A U-Net is
convolutional, local and multiscale, PDEBench's own baseline alongside the Fourier operator, and
the first architecture here that is not global. It learns both families: against the identity map
it is better by $16.9\times$ on advection and $491.9\times$ on reaction--diffusion, and against
predicting the training mean by $23.8\times$ and $58.7\times$. It is nonetheless the weaker
operator on both, at $4.22\%$ against the Fourier operator's $0.857\%$ on advection and $0.116\%$
against $0.0972\%$ on reaction--diffusion, factors of $4.9$ and $1.2$. The gap on advection is
what a stack of local convolutions gives up to a spectral method on a pure shift, which a
truncated Fourier transform represents exactly. Both families clear the feasibility gate, at
$2.54\times$ and $4.64\times$ at the lowest rate.

Three conditions were fixed in advance and two of them were missed.

\paragraph{Calibration.} We required the worst ratio of predicted to measured cost
over the U-Net points to stay at or under the $2.11\times$ the paper quotes over six families and
two architectures. It reached $\mathbf{2.85\times}$, on reaction--diffusion at a sixteenth of a
bit per value, where the probe reads $13.21\times$ against a measured $4.64 \pm 0.52\times$. The
figure has now moved with every family we have added and with the one architecture that does not
mix globally. \textbf{We therefore stop quoting a single calibration number.} What survives is
the ordering, and the interval on the worst cell, not the point estimate.

\paragraph{Ordering.} The probe's rank correlation with the measured cost
over the eight U-Net points is $\mathbf{+0.952}$ against a pre-registered $+0.70$.

\paragraph{Decoupling.} We required the probe
to invert strictly fewer of the 16 across-family pairs than field error. Both invert
\emph{one}. On two families whose cost orderings happen to agree with their field errors there is
nothing for the test to separate, and this arm therefore neither supports nor contradicts
Section~\ref{sec:decouple}, whose claim rests on six families and 104 pairs. We
record it as a missed bar rather than as a null result, because it is the bar we wrote. One
further thing did move: field error inverts 1 of 12 \emph{within}-family pairs here, the
first time that has happened in this paper, because two reaction--diffusion cells lie at
$0.98\times$ where training on compressed data beat full precision inside the seed spread. The
within-family monotonicity is a property of the sweep, and at the noise floor it stops holding.

Three architectures, then, and what we can say is narrower than what two suggested: the ordering
holds on all three, the magnitude does not transfer, and the decoupling has been measured on the
two global ones.

\section{The excluded families}
\label{app:excluded}

Section~\ref{sec:probe} states the gate and the verdict. These are the two arms it excluded and
the one we rejected before running.

PDEBench 2D diffusion--reaction draws its initial condition as noise. Its field error is
$91\%$ at a quarter bit per value, where every family we do report is under $15\%$, and it needs
4 bits to reach $10\%$, which is eight-fold compression against the $128$ to $263$-fold the others reach. A flat spectrum leaves a transform codec nothing to remove. The full-precision model
reached $98\%$ error and every measured cost was $1.00\times$. Taking the velocity channel of 1D compressible CFD failed the
gate for a different reason: one field of a compressible system is not
a closed map, the full-precision error was $78\%$, and again every cost was $1.00\times$. Neither arm constrains the probe in either direction, and counting them as passes would have
inflated the result. We also rejected 2D incompressible Navier--Stokes before running it, because the file
holds four long trajectories rather than an ensemble of initial conditions and cannot be split the
way the rest are.

The practical point for a curator is that the premise of this paper, that PDE fields are smooth
and redundant, is a property of particular stored datasets and not of PDEs. It should be verified on the field in question before a compression budget is planned around it.

\section{The candidate rate checked against training}
\label{app:rstarcheck}

\begin{figure}[!ht]
\centering
\includegraphics{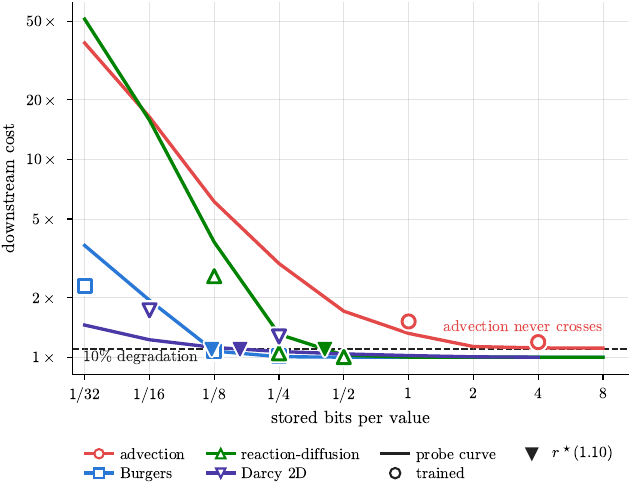}
\caption{Downstream cost against stored bits per value, four families. Lines are probe curves,
hollow marks are trained measurements, and triangles mark $r^\star(1.10)$.}
\label{fig:rstar}
\end{figure}

Table~\ref{tab:rstarcheck} sets the candidate rate nominated by the probe against every trained
cell of the same family, all on the corrected codebook, and Figure~\ref{fig:rstar} shows the
curves the rate is read off.

\begin{table}[!ht]
\centering
\footnotesize
\begin{tabular}{lrrrl}
\toprule
family & probe $r^\star$ & trained fit & ratio & the check \\
\midrule
Burgers & $0.1217$ & $0.1037$ & $0.85$ & $0.125$ bpv costs $1.07\times$, consistent \\
reaction--diffusion & $0.4080$ & $0.2234$ & $0.55$ & $0.25$ bpv, below it, costs $1.05\times$: \textbf{conservative} \\
Darcy 2D & $0.1647$ & --- & --- & $0.25$ bpv costs $\mathbf{1.27\times}$, \textbf{outside} \\
shallow water 2D & $3.349$ & $6.836$ & $2.04$ & extrapolated, directional only \\
reaction--diffusion, far & $0.4130$ & $0.7006$ & $1.70$ & $0.5$ bpv costs $\mathbf{1.15\times}$, \textbf{outside} \\
advection & $>\!8$ & --- & --- & $1$ and $4$ bpv cost $1.52$ and $1.19\times$, consistent \\
\bottomrule
\end{tabular}
\caption{The back-check. The trained fit is where a line through $\log(\mathrm{cost}-1)$ against
$\log r$ over a family's trained cells crosses the tolerance; the ratio is that fit over the
probe's rate.}
\label{tab:rstarcheck}
\end{table}
\FloatBarrier
\section{What the probe saves}
\label{app:cost}
 A training run on the hardware used here takes
$404$ seconds, the median over the 28 training runs of our first rate sweep. Producing an
eight-rate curve for one family by training at every rate, with three seeds, is 24 runs,
or $2.69$ GPU-hours. The probe curve alone is three runs, $0.34$ GPU-hours, a factor of $8.0$;
with two rates additionally validated by training it is 9 runs, $1.01$ GPU-hours, a factor of
$2.7$. Over its four families, training at every rate would have cost $10.8$ GPU-hours against the
$4.0$ that sweep spent. The forward passes that produce the curve are
not separated out in these figures because they fall below the resolution of the log timestamps:
all eight rates are evaluated within the same second as the training run that precedes them.

\section{The codec comparison in full}
\label{app:ratefair}

Section~\ref{sec:ratefair} states four results from this comparison; this appendix gives the
arms it ran, the tables behind them and the corrections it produced. Writing $\lvert
C(\cdot)\rvert$ for the length in bits of a codec's output, the two accountings are
\begin{equation} r_{\mathrm{ps}} \;=\; \frac{1}{nD}\sum_{i=1}^{n} \bigl\lvert C(A_i) \bigr\rvert
, \qquad r_{\mathrm{ens}} \;=\; \frac{1}{nD} \bigl\lvert C(A_1, \dots, A_n) \bigr\rvert ,
\label{eq:units} \end{equation}

The compressors are the shipping implementations, not our reconstruction of them: SZ3
\citep{liang2022sz3} through the SZ3 team's own bindings with the multi-level interpolation
predictor and the Huffman-plus-Zstd back end, and ZFP \citep{lindstrom2014fixed} through
\texttt{zfpy} in its fixed-accuracy mode, which is the mode that reaches our rates. Its
fixed-rate mode does not: on a $1024$-point field it faults rather than returns for every rate
we asked for below 4 bits per value, and on the $64 \times 64$ Darcy fields below one, where its
reconstruction is worse than a field of zeros. Both are swept from a sixteenth of a bit per
value to eight, so the comparison spans the range a computing facility works in as well as the
range operator learning can live in. Section~\ref{sec:limitations} records what changed when we
replaced the stand-in we had written, since the correction counts against us.

\paragraph{Side information is now charged.} Equation~\eqref{eq:side} is $0.0068$ bits per value
here, 11 per cent of a $0.0625$ budget, which Section~\ref{sec:accounting}'s own rule says must
be paid for and which earlier versions did not. Every rate below is the total, and the payload
is reduced to make room. This is why our reported numbers
have moved: advection at $0.0625$ bits per value was reported at $14.63\%$ field error and
$16.46\times$ cost, and paying for the side information it costs $17.62\%$ and
$\mathbf{19.48\times}$.

\begin{table}[!ht]
\centering
\small
\begin{tabular}{llrrr}
\toprule
family & arm & rate & field error & cost \\
\midrule
advection & water-filling             & $0.0625$ bpv & $17.62\%$  & $\mathbf{19.48 \pm 0.76\times}$ \\
advection & best low-pass             & $0.0625$ bpv & $16.38\%$  & $20.11 \pm 0.34\times$ \\
advection & SZ3, whole ensemble       & $0.0635$ bpv & $151.42\%$ & $95.37 \pm 1.54\times$ \\
advection & SZ3, per sample, at its floor & $0.8984$ bpv & $100.00\%$ & $116.78 \pm 0.01\times$ \\
advection & ZFP, per sample, at its floor & $0.3750$ bpv & $100.00\%$ & $116.78 \pm 0.01\times$ \\
\midrule
reaction--diffusion & best low-pass       & $0.0625$ bpv & $10.97\%$  & $\mathbf{10.99 \pm 0.87\times}$ \\
reaction--diffusion & water-filling       & $0.0625$ bpv & $11.82\%$  & $12.11 \pm 1.62\times$ \\
reaction--diffusion & SZ3, whole ensemble & $0.1020$ bpv & $60.09\%$  & $56.45 \pm 1.36\times$ \\
reaction--diffusion & SZ3, per sample, at its floor & $0.8984$ bpv & $100.00\%$ & $69.81 \pm 0.04\times$ \\
reaction--diffusion & ZFP, per sample, at its floor & $0.3750$ bpv & $100.00\%$ & $69.81 \pm 0.04\times$ \\
\bottomrule
\end{tabular}
\caption{A budget of $64$ bits per sample spent five ways, against the shipping SZ3 and ZFP,
over two seeds. Rates are totals including side information, and the two per-sample rows sit at
their floors because no error bound brings them to the budget.}
\label{tab:ratefair}
\end{table}

\paragraph{Per-sample rate floors.} Compressing
each field on its own, no error bound takes SZ3 below $\mathbf{0.898}$ bits per value on a
$1024$-point field or ZFP below $\mathbf{0.375}$, against the $0.0625$ this paper operates at:
factors of $14.4$ and $6.0$. On the $64 \times 64$ Darcy fields the floors are $0.230$ and
$0.094$. The reason is structural. A compressed unit carries a header, an exponent and at least one coded
plane, a fixed cost $H$ that does not depend on the field, so the per-sample rate cannot fall
below $H/D$. The four floors say exactly that: $H$ is $920$ and $944$ bits for SZ3 across a
fourfold change in $D$, and $384$ bits for ZFP at both sizes.
At the floor both return relative errors of exactly $100\%$, the value a coder returns when it
has kept nothing. Compressing the whole ensemble as one array removes the floor entirely,
taking SZ3 to $0.0001$ bits per value, because the header is then amortised over ten thousand
fields; it also removes the ability to read one field without the rest. \textbf{The operating
point this paper studies is outside the range in which either compressor can deliver an
independently decodable field at all.} This answers a question posed in advance, and is the more consequential half of what this
section measures.

\begin{figure}[!ht]
\centering
\includegraphics{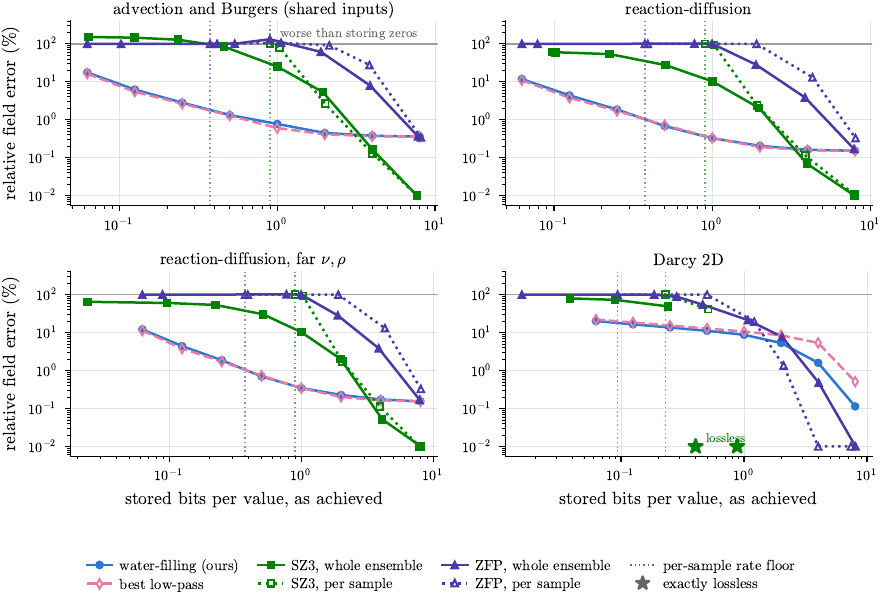}
\caption{Rate against distortion over seven doublings, four input ensembles. Curves stop at each
per-sample arm's floor, marked by the dotted vertical lines. On Darcy, SZ3 becomes exactly
lossless at $0.404$ bits per value, drawn as a star at the axis floor.}
\label{fig:ratedistortion}
\end{figure}

\paragraph{The crossover, which is the useful answer.} Figure~\ref{fig:ratedistortion} shows the
whole range, and it does not have one winner. Below 2 bits per value spectral water-filling is
far ahead: at a quarter of a bit it beats real SZ3 by $\mathbf{45.6\times}$ on advection, by
$28.9\times$ on reaction--diffusion and by $28.7\times$ at the far parameter point. Between 2
and 4 bits per value that reverses on every one-dimensional family, and by 8 bits SZ3 is
ahead by more than two orders of magnitude, at $0.0019\%$ against our $0.3581\%$. Our codec has a
floor of its own that SZ3 does not: a cap of 14 bits per coefficient and a fixed transform
leave advection stuck at $0.358\%$ however many bits it is given. \textbf{The practical statement
is therefore a range and not a ranking.} For operator learning, where the interesting rates are
below a quarter of a bit, use a spectrum-adapted transform code. For archival storage at the 2
to 8 bits a facility uses, use SZ3.

\paragraph{Darcy.} PDEBench draws the Darcy permeability field from two
values, and real SZ3 compresses it \emph{exactly}, with zero error, at $0.404$ bits per value.
Our codec at $0.5$ bits per value still has $11.17\%$ field error, and at $8$ bits per value,
20 times the storage, it has $0.11\%$ and has still not reached zero. A discrete-cosine basis
is the wrong representation for a piecewise-constant field and Lorenzo prediction followed by
Zstd is the right one. For that family the recommendation is to store it losslessly with SZ3 at four tenths of a bit
per value, which answers this paper's question for that dataset better than anything else in it: its
training inputs need $0.404$ bits per value and no approximation at all.

\paragraph{The stand-in was generous, not pessimistic.} An earlier
caption here said that pricing the stand-in's index stream at order-$0$ empirical entropy was
pessimistic, because SZ3 follows Huffman with Zstd and Zstd exploits the runs that stream is
mostly made of. Measured, the opposite holds. At the same rate the real SZ3 reaches $151.4\%$
field error on advection and $60.1\%$ on reaction--diffusion, against the stand-in's $57.1\%$ and
$53.4\%$; running SZ3's Lorenzo predictor alone, the closest match to what we had written, gives
$116.4\%$ and $54.1\%$. The order-$0$ pricing charged nothing for headers, block metadata or the
predictor's own state, and at a sixteenth of a bit per value those are the entire budget. The stand-in therefore credited SZ3 with a rate-distortion point it does not achieve. This caption had already been corrected once, in the wrong direction; both corrections are
reported.

\paragraph{Replacing the flat-allocation arm.} Spreading 1 bit over the
top 64 modes gave $95.41\%$ field error, which is not a configuration any practitioner
would choose. The baseline a practitioner would actually reach for is a low-pass truncation: retain the $K =
\lfloor B/b \rfloor$ highest-variance coefficients at a uniform $b$ bits, and take the $b$ that
minimises field error at the same budget $B$. It is a real competitor. It beats water-filling on field error at almost every
rate we tested, and downstream the two are within the seed spread of each other, at $19.48$
against $20.11\times$ on advection and $12.11$ against $10.99\times$ on reaction--diffusion.
\textbf{Water-filling's advantage over a well-chosen low-pass truncation is not established by
our measurements}, and the argument of this section rests on the transform, not on the
allocation rule inside it.

\paragraph{A reversed conclusion.} An earlier comparison matched field error rather than rate
and found the opposite ordering between the transform code and the error-bounded one; the
accounting rule of Section~\ref{sec:accounting} states the confound. Two of the four arms in
that comparison were not codecs, being an additive perturbation and a projection retained at
full precision, so they had no rate to report at all.

Note finally what field error does inside Table~\ref{tab:ratefair}. On reaction--diffusion it
orders all 10 pairs correctly. On advection it inverts 3 of 10, because the two arms with
$100\%$ error cost less downstream than the ensemble arm at $151\%$, and because the low-pass arm
has the lower field error and the higher cost. The decoupling of Section~\ref{sec:decouple} therefore appears here too, at fixed rate and
within a single family. An earlier version of this paragraph said that it did not.

\section{Spending a budget on samples or on precision}
\label{app:frontier}

Two budgets, a quarter of a bit and 1 bit per value over the full training pool, on the two
families that lie at opposite ends of this paper's cost range. Charging the side information of
Equation~\eqref{eq:side} against a budget of $S$ bits, $n$ samples of $D$ values at payload rate
$r$ occupy $D\,(nr + 2b_{\mathrm{f}} + b_{\mathrm{a}})$ bits, so the budget line and the point
we report on it are \begin{equation} r(n) \;=\; \frac{1}{n}\left( \frac{S}{D} - 2b_{\mathrm{f}}
- b_{\mathrm{a}} \right) , \qquad n^\star \;=\; \arg\min_{n} \; \varepsilon\bigl(n, r(n)\bigr) .
\label{eq:line} \end{equation} A feasibility gate was registered in advance: $\varepsilon(n_{\min}, \infty) /
\varepsilon(n_{\max}, \infty) \ge 1.10$, or samples do not matter in this range and the line
cannot separate the two ways of spending. All four cases pass it, at $2.18$,
$1.25$, $2.33$ and $1.41\times$.

\begin{figure}[!ht]
\centering
\includegraphics{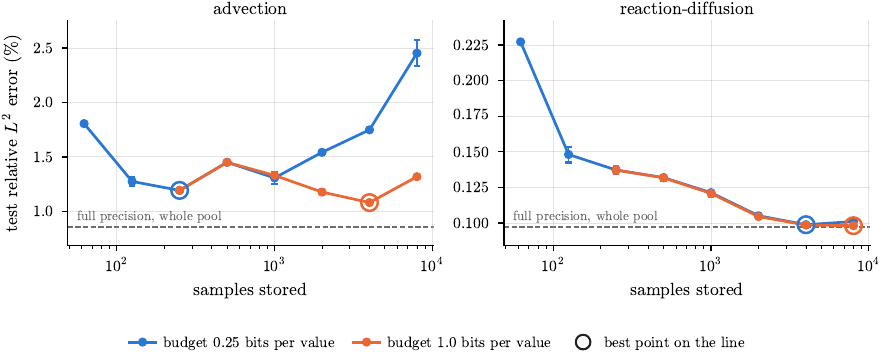}
\caption{Every point on a curve costs the same number of bytes: right stores more fields at
lower precision, left fewer at higher. The ring marks the best point on each line and the dashed
line a model trained on the whole pool. Two seeds, error bars are their spread.}
\label{fig:frontier}
\end{figure}

\begin{table}[!ht]
\centering
\small
\begin{tabular}{llrrrr}
\toprule
family & budget & best point & its error & whole pool & gain \\
\midrule
advection & $0.25$ bpv & $250$ at $7.73$ bpv & $\mathbf{1.192\%}$ & $2.454\%$ & $\mathbf{2.06\times}$ \\
advection & $1.0$ bpv  & $4000$ at $1.98$ bpv & $\mathbf{1.080\%}$ & $1.318\%$ & $1.22\times$ \\
reaction--diffusion & $0.25$ bpv & $4000$ at $0.48$ bpv & $0.0987\%$ & $0.1010\%$ & $1.02\times$ \\
reaction--diffusion & $1.0$ bpv  & $8000$ at $0.99$ bpv & $0.0981\%$ & $0.0981\%$ & $1.00\times$ \\
\bottomrule
\end{tabular}
\caption{The best point on each budget line against the point that keeps every sample, which is
the configuration the rest of this paper uses. Two seeds.}
\label{tab:frontier}
\end{table}

\paragraph{The optimum is family-dependent.}
Figure~\ref{fig:frontier} draws the four budget lines and Table~\ref{tab:frontier} gives the best
point on each. We pre-registered that the best point would be at the largest sample count, as operator learning
being sample-hungry predicts and as this paper implicitly assumes every time it recommends compressing rather than subsampling. \textbf{That failed on 3 of the
4 budget lines.} On advection at a quarter of a bit per value the optimum stores 250 fields at high precision
rather than the full pool at low precision, and beats it by a factor of $2.06$; the full pool is
the worst point on that line. Against a
full-precision model on the whole pool the best point costs $1.39\times$ where our configuration
costs $2.86\times$. At the wider budget of 1 bit per value the optimum moves back to four
thousand samples and the line flattens to a $1.34\times$ spread, so the choice stops mattering.
On reaction--diffusion the optimum is at 4000 and 8000 samples, the largest or
next to largest, and the curve degrades monotonically as samples are given up.

\paragraph{Agreement with the transmission coefficient.} Advection transmits
essentially all of its codec error to the output, at $0.97$ to $1.12$ in
Table~\ref{tab:transmit}, so precision is costly to give up there and the budget is better spent on it. Reaction--diffusion transmits $0.032$ to $0.084$, so precision is nearly free to
give up and samples are what it lacks. The same forward-pass quantity that orders downstream
cost in Section~\ref{sec:decouple} therefore also says which way to spend a storage budget, a
second use for it that we did not anticipate. Two families cannot establish that, and we
state it as the reading of these four lines rather than as a result. What the four lines do
establish is narrower and still useful: \textbf{on at least one PDE family, the configuration
this paper has been recommending is the worst way to spend its own budget.}

\section{What the codec stores against what the operator pays}
\label{app:floor}

\begin{figure}[!ht]
\centering
\includegraphics{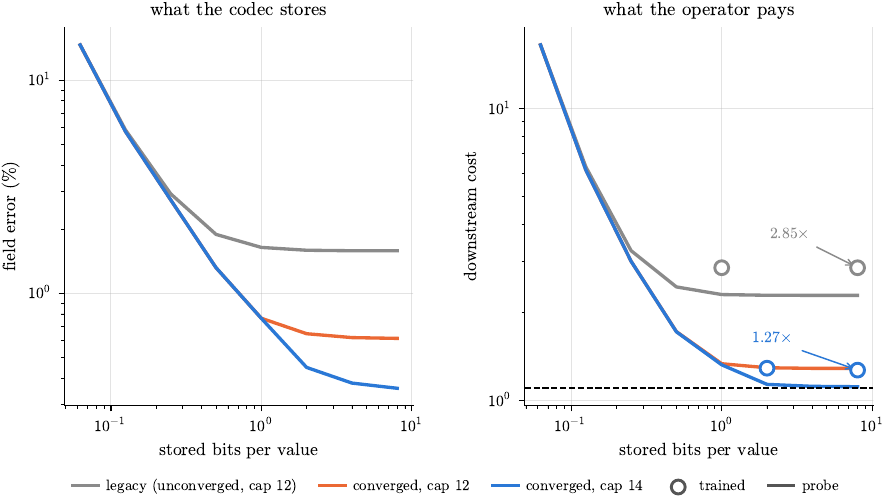}
\caption{Advection. Left, what the codec stores; right, what the operator pays, under the legacy
unconverged Lloyd table and the converged companded codebook at caps of $12$ and $14$ bits.
Lines are probe predictions, hollow circles trained measurements, and the dashed line the $10\%$
threshold.}
\label{fig:floor}
\end{figure}

Section~\ref{sec:codec} states the correction. The two panels of Figure~\ref{fig:floor} separate
what the codec stores from what the operator pays, so that the plateau can be seen to be a
property of the codebook and not of the field.

\section{The seven explanations in full}
\label{app:refuted}

Each of these is refuted in Section~\ref{sec:refuted}; this appendix gives the
measurement behind every verdict.

\paragraph{Support dimension.} A projection-type codec confines the training inputs to a fixed
subspace, so the clean test input lies off it and the regime becomes one of distribution shift.
We tested this by filling the discarded coefficients with draws from their own marginal, seeded
so the decoder reproduces them at zero bit cost. The hard support rank of the reconstruction rose
from $17$ to $1024$ as designed, and the downstream cost changed by $1.03\times$ and
$0.98\times$, that is, not at all. The support-geometry explanation is wrong; the filled
directions carry no information about the target, so the model learns to ignore them.

\paragraph{Information spreading.} If what matters is the number of coefficients carrying
non-zero information, then spreading a fixed budget over more of them should help. Three arms at
$64$ bits per sample now put a number on that. Three arms at the same budget inform $64$, $17$ and $11$ modes respectively, and the cost does
not follow that count in either direction: the arm that spreads most is far the worst, while the
two that concentrate are indistinguishable within the seed spread. Table~\ref{tab:ratefair}
gives the values. A randomised allocation,
intended to make the error sample-dependent at zero side-information cost, also failed, and
diagnostically it never moved its own independent variable: the effective rank of the error matrix
went from $10.0$ to $3.7$, in the wrong direction.

\paragraph{Operator-aware bit allocation.} Perturbing transform coefficient $i$ moves the output
along $W c_i$, suggesting water-filling on $\sigma_i^2\lVert Wc_i\rVert^2$. We built in a
falsifier: on advection, whose solution operator is a shift, the two weightings must coincide.
The falsifier passed, at $1.000$, $1.004$, $0.997$ and $1.000$. Elsewhere the gains are
non-monotone in rate and the method loses at $1$ bit per value, so we report it as a negative.

\paragraph{A heavy-tailed codebook.} The standardised transform coefficients have live-mode
excess kurtosis of $117$ and $209$. A generalised Gaussian whose shape parameter is clamped at $0.5$, below which the density is
numerically unusable on a bounded grid, expresses an excess kurtosis of at most $22.2$. The family cannot represent these tails, so this fix needs a different codebook
family rather than a better fit to the same one.

\paragraph{Errors adding in quadrature.} Write $G$ for the model trained on $\hat A = A +
\delta$ and $F^{*}$ for the truth. Expanding $G$ about $\hat A$ at $A = \hat A - \delta$,
\begin{equation} G(A) - F^{*}(A) \;=\; \zeta(\hat A) \;-\; DG[\delta] \;+\;
O\bigl(\lVert\delta\rVert^{2}\bigr) , \qquad \zeta(\hat A) \;=\; G(\hat A) - F^{*}(A) ,
\label{eq:quad} \end{equation} and taking norms with the two terms uncorrelated gives
Equation~\eqref{eq:pred}. Measured directly they are not: the required assumption that $\zeta$
matches $\varepsilon_0$ fails by up to $12.9\times$, and the implied correlation between
residual and displacement is $-0.62$ to $-0.97$ rather than zero. The probe's form works empirically while its natural derivation does not.

\paragraph{Codec error outside the band the operator reads.} The one explanation here that came
from a measurement rather than from intuition, and the one that came closest. Section~\ref{sec:probe}
shows our codec confining its error to the coefficients that carry signal while a per-sample
bounded-error codec, here the stand-in of Table~\ref{tab:sz3} rather than the shipping SZ3,
spreads a residual over all of them, and the probe is accurate for the first and over-predicts
for the second and for the wavelet. Since the spectral convolutions keep only 16 modes, error above
that band reaches the output through the pointwise paths, where a forward pass counts it and a
model trained on the same data can learn to discount it. That would make the probe read high by
the amount of error the operator does not need.

We pre-registered two conditions. The fraction of codec error energy above mode $16$ should
rank-correlate with the over-prediction at $\rho \ge 0.6$; and, because three codecs make any
statistic that separates them correlate with anything else that separates them, the same effect
had to appear \emph{within} a codec as the rate varies. Measured on $2000$ samples per family
with all three codecs matched to the same field error, the out-of-band fraction is $7.7\%$ for
our codec, $22.8\%$ for the wavelet and $85.6\%$ for SZ3, in the same order as the
over-predictions of $1.00$, $3.49$ and $4.65\times$. The first condition passes at
$\rho = +0.826$. Inside the wavelet arm, the only one with four rates on two families, the
correlation is $\mathbf{-0.667}$. The sign reverses, so the quantity is telling us which codec we
are in rather than why it behaves as it does, and the second condition is what says so. We report
it as refuted. A first pass at this measurement used the three sample fields of Figure~\ref{fig:errstruct} and
produced the reverse ordering, because the allocation was water-filled on the variance of the
same three samples it then compressed; that is why it was repeated at scale rather than
reported.

\paragraph{Attenuation bias, the model behind the verdict.} Section~\ref{sec:refuted} states the
objection and the outcome. Write the observed log field error as $\log \tilde e = \log e + \eta$
with $\eta \sim \mathcal{N}(0, \sigma^2)$ independent across points. Two points whose true log
field errors differ by $\Delta$ are then inverted with probability \begin{equation} \pi(\Delta;
\sigma) \;=\; \Pr[\,\text{inversion} \mid \Delta\,] \;=\;
\Phi\!\left(\frac{-\Delta}{\sigma\sqrt{2}}\right) , \label{eq:eiv} \end{equation} so any set of
pairs with known separations $\Delta_k$ and inversion indicators $z_k$ gives \begin{equation}
\ell(\sigma) \;=\; \sum_k \Bigl[\, z_k \log \pi(\Delta_k; \sigma) \;+\; (1 - z_k) \log\bigl(1 -
\pi(\Delta_k; \sigma)\bigr) \Bigr] , \label{eq:eivfit} \end{equation} maximised at $\hat\sigma$,
with a one-sided $95\%$ upper bound $\sigma_{95} = \max\{\sigma : \ell(\sigma) \ge
\ell(\hat\sigma) - \tfrac{1}{2}\chi^2_{1,0.90}\}$ by the likelihood ratio.

Two conditions were fixed before either was computed. The $\sigma$ explaining the across-family
inversions must be consistent with what the 16 within-family pairs permit, those being the same
quantity measured by the same code on the same fields. The inverted pairs must also concentrate
where the two points are close together, since the first condition alone can be satisfied by a
hypothesis about separation rather than about noise.

\section{Decimation against area averaging}
\label{app:grid}

Section~\ref{sec:limitations} states the verdict. Area averaging is the $2 \times 2$ block mean,
the $L^2$ projection onto the coarse grid, which does not alias where decimation does.

Three conditions were fixed in advance. The fitted spectral exponent moves by $5.4\%$, from
$2.78$ to $2.93$, against a bar of $10\%$. That exponent is the radial decay of the
transform-coefficient variance, which in two dimensions runs about a factor of $d$ above the
index-ordered eigenvalue exponent $\gamma_{\text{in}}$ of Appendices~\ref{app:spectral}
and~\ref{app:structure}; it is the same field measured a different way, and what matters here is that both restrictions
are measured the same way.

Field error at a quarter of a bit moves by $17.5\%$, from
$13.54\%$ to $11.18\%$, against a bar of $20\%$; area averaging is the more compressible of the
two, as it should be. The downstream cost that the tables report agrees to within $\mathbf{1.08\times}$: $1.727$ against $1.667\times$ at a sixteenth of a bit and $1.273$
against $1.175\times$ at a quarter. The run also reproduced this paper's own Darcy numbers from
weeks earlier, $13.540\%$ against $13.539\%$ and $19.565\%$ against $19.562\%$. The restriction
was harmless; had it not been, the two-dimensional rows would have been requoted under area
averaging.

\section{The pre-registered bars}
\label{app:bars}

The two attenuation rows are scored on the single-protocol points; on the earlier two-quantiser
assembly they read $2.71$ and $1.15$ against $1.74$, with the same verdicts.
Table~\ref{tab:bars} lists all thirty-eight, with the value each would have had to reach to
fail.

\begin{table}[!ht]
\centering
\scriptsize
\begin{tabular}{llll}
\toprule
bar & required & would have failed at & outcome \\
\midrule
geometric mean, out of sample & $\le 1.30\times$ & any worse & \textbf{MISSED}, $1.43\times$ \\
probe inside SZ3 & $\le 2.0\times$ & any worse & \textbf{MISSED}, $4.65\times$ \\
probe inside a wavelet, worst & $\le 1.61\times$ & any worse & \textbf{MISSED}, $3.49\times$ \\
probe inside a wavelet, median & $\le 1.25\times$ & any worse & \textbf{MISSED}, $2.14\times$ \\
2D Fourier operator under-predicts & $\ge 3$ of $4$ & $\le 2$ of $4$ & \textbf{MISSED}, $0$ of $4$ \\
worst case, sixth family & $\le 2.10\times$ & any worse & \textbf{MISSED}, $2.11\times$ \\
band content, within a codec & $\rho \ge 0$ & any negative & \textbf{MISSED}, $-0.667$ \\
attenuation, noise consistency & $\sigma \le 0.16$ & any larger & \textbf{MISSED}, $2.38$ \\
real SZ3 beats our stand-in & $\le 57.1\%$ & any worse & \textbf{MISSED}, $151.4\%$ \\
budget optimum at the most samples & all $4$ lines & any fewer & \textbf{MISSED}, $1$ of $4$ \\
worst case, third architecture & $\le 2.11\times$ & any worse & \textbf{MISSED}, $2.85\times$ \\
decoupling, third architecture & probe inverts fewer & a tie or worse & \textbf{MISSED}, $1$ vs $1$ \\
single protocol, field errors reproduce & within $15\%$ & any wider & \textbf{MISSED}, $50.9\%$ \\
relative $L^2$, normalised, separation & $\ge 2\times$, $\le 1$ block & either breached & \textbf{MISSED}, $1.2\times$, $0$ \\
relative $L^2$, physical, separation & $\ge 2\times$, $\le 1$ block & either breached & \textbf{MISSED}, $1.8\times$, $0$ \\
\midrule
ordering, fifth family & $\rho \ge +0.60$ & \emph{unreachable}\footnotemark[1] & met, $+0.886$ \\
ordering, sixth family & $\rho \ge +0.70$ & \emph{unreachable}\footnotemark[1] & met, $+0.853$ \\
band content, between codecs & $\rho \ge +0.60$ & \emph{unreachable}\footnotemark[2] & met, $+0.826$ \\
attenuation, inversions are near-ties & median below & any above & met, $1.15$ vs $1.80$ \\
ours below real SZ3 at $0.0625$ bpv & both families & either lost & met, $8.6\times$, $5.1\times$ \\
ours below real SZ3 downstream & both families & either lost & met, $19.5$ vs $95.4\times$ \\
SZ3 ahead somewhere at $4$ bpv & $\ge 1$ family & none & met, $4$ of $4$ \\
neither codec reaches $0.25$ bpv & per sample, 1D & either did & met, $0.898$, $0.375$ \\
budget line spread & $\ge 1.5\times$ once & all flatter & met, $2.06\times$, $2.30\times$ \\
restriction, spectral exponent & $\le 10\%$ apart & any wider & met, $5.4\%$ \\
restriction, field error & $\le 20\%$ apart & any wider & met, $17.5\%$ \\
restriction, downstream cost & $\le 1.25\times$ & any wider & met, $1.08\times$ \\
ordering, third architecture & $\rho \ge +0.70$ & any lower & met, $+0.952$ \\
single protocol, separation & $\ge 2\times$, $\le 1$ block & either breached & met, $3.0\times$, $0$ \\
single protocol, fp32 errors reproduce & within $10\%$ & any wider & met, $0.34\%$ \\
relative $L^2$, normalised, $\kappa$ order & $\rho \ge 0.9$ & any lower & met, $1.00$ \\
relative $L^2$, normalised, consistency & fields, fp32 & either wider & met, $0.000\%$, $0.0\%$ \\
relative $L^2$, physical, $\kappa$ order & $\rho \ge 0.9$ & any lower & met, $1.00$ \\
relative $L^2$, physical, consistency & fields, fp32 & either wider & met, $0.000\%$, $3.9\%$ \\
relative $L^2$, physical, stable seeds & both within $15\%$ & either wider & met, $1.1\%$, $2.1\%$ \\
\midrule
DeepONet does not under-predict & $\ge 2$ of $3$ & \emph{unreachable}\footnotemark[3] & void \\
worst case, diffusion--reaction & $\le 2.10\times$ & --- & void, gate not met \\
ordering, diffusion--reaction & $\rho \ge +0.70$ & --- & void, gate not met \\
\bottomrule
\end{tabular}
\footnotetext[1]{The bar sat below the value already in hand, and the run added one family whose
four points are monotone in both variables by construction, so the pooled correlation was pinned
by the points already there.} \footnotetext[2]{The dependent variable takes one value per codec,
so this is a three-item ordering whose smallest attainable $p$ is $1/6$. It is why the
within-codec condition was pre-registered alongside it, and that one did the work.}
\footnotetext[3]{The probe is $\sqrt{1+(T/\varepsilon_0)^2} \ge 1$ identically, while a measured
cost on a dead arm dips below $1$ from seed noise, so the ratio cannot fall short.}

\caption{Every bar written into a run script before the run that scored it. Fifteen of the
thirty-eight were missed, twenty were met, and three are marked void because they could not have
failed.}
\label{tab:bars}
\end{table}
\section{The second predictor}
\label{app:tiers}

The probe has a second error pattern, along a different axis from the codec dependence just
described. Holding the codec fixed and varying the family, it \emph{under}-predicts on advection
and on 2D Darcy while being accurate on reaction--diffusion and Burgers, whereas the SZ3 result
above is an \emph{over}-prediction driven by the codec. The two are separate and we treat them
separately. Under-prediction means a model
\emph{trained} on compressed data is worse than a clean model merely \emph{evaluated} on it,
which a forward pass cannot see by construction.

We therefore added a second predictor: fine-tune the full-precision model on the compressed data
for $15$ epochs, a tenth of a training run, and divide its error by that of the same model
fine-tuned for the same $15$ epochs on \emph{clean} data. The control matters. Without it the
tier is credited with $15$ epochs of ordinary optimisation, and on an undertrained model it
appears to beat full precision for that reason alone. Measured control ratios were $0.981$ to
$1.009$.

On 8 measurements spanning two families and two architectures, the probe read high in five
cells and the fine-tune low in six, so the two bracketed the truth in 5 of the 8, and
their geometric mean was much tighter than either: worst case $1.14\times$ against $1.49\times$
and $1.36\times$.

That rule was chosen after seeing those eight numbers and would have been scored on the same
eight. We therefore froze it, with no remaining freedom, and pre-registered a bar in the run
script before executing it: the geometric mean should come in under $1.30\times$ on families
that took no part in choosing it, and should beat both individual tiers.

\begin{table}[!ht]
\centering
\small
\begin{tabular}{lrrr}
\toprule
predictor & in-sample (8) & out-of-sample (8) & mean $|\log|$, out of sample \\
\midrule
forward-pass probe        & $1.49\times$ & $1.61\times$ & $0.164$ \\
$15$-epoch fine-tune      & $1.36\times$ & $1.66\times$ & $0.165$ \\
geometric mean of the two & $1.14\times$ & $\mathbf{1.43\times}$ & $0.141$ \\
\bottomrule
\end{tabular}
\caption{Worst-case factor off. The pre-registered bar was $1.30\times$ out of sample. It was
not met.}
\label{tab:tiers}
\end{table}

\textbf{The bar was not met}, at $1.43\times$. The bracketing itself held up, in 4 of 8
cells out of sample against 5 of 8 in sample, so the structure the rule was built on
survived and its accuracy did not. The in-sample $1.14\times$ was optimism of exactly the kind
the pre-registration was written to catch. The geometric mean remains the best of the three tiers
out of sample, and pooled over all $16$ points it gives $1.43\times$ against $1.61\times$ and $1.66\times$, but the figure we report is $1.4$ to $1.6$, not $1.14$.

\begin{figure}[!ht]
\centering
\includegraphics{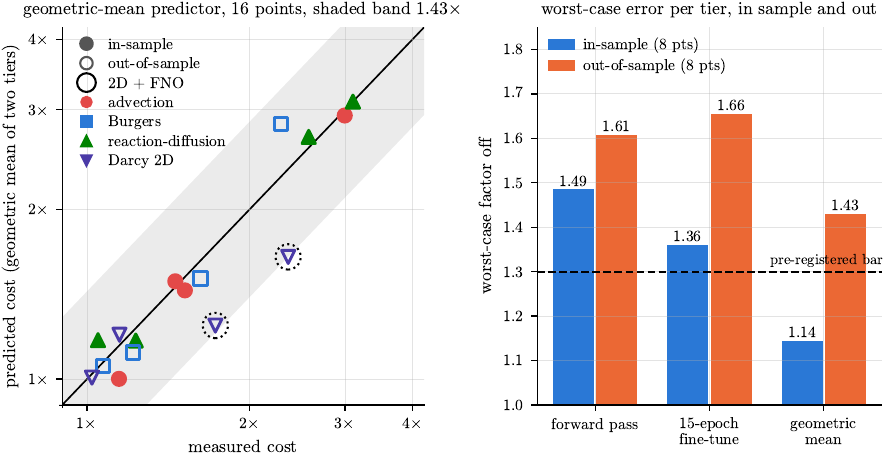}
\caption{Left: the geometric-mean predictor on all $16$ points, in-sample filled, out-of-sample
open, the 2D Fourier neural operator cells ringed. Right: worst-case error per tier against the
pre-registered bar.}
\label{fig:tiers}
\end{figure}

Figure~\ref{fig:tiers} shows both halves of this: the predictions against measurement on the
left, and on the right the three tiers degrading from in-sample to out-of-sample against the bar.

\paragraph{Localising the failure.} Excluding the two-dimensional Fourier
neural operator, the geometric mean over the remaining $14$ points is $\mathbf{1.24\times}$.
Every tier under-predicts that one cell, and the probe has done so at ratios $0.74$, $0.86$,
$0.81$ and $0.71$, spread over three separately executed runs at the same fixed seeds, so these
are re-executions rather than fresh draws. Four measurements in the same
direction is a reproducible failure mode of the method, not noise, and we report it as an open
limitation rather than averaging it away.

The failure is not that two-dimensional data is harder to predict. The same Darcy dataset under
DeepONet is predicted to $1.13\times$ and $0.98\times$, while under the Fourier neural operator
it gives $0.81\times$ and $0.71\times$. Holding the data fixed and changing only the architecture
removes the effect.

Nor is the failure a property of the Fourier neural operator in two dimensions: we
pre-registered that hypothesis and it failed. On 2D shallow water we required the probe to
under-predict at 3 of the 4 trained rates; it \emph{over}-predicted at all four, by $1.62$ to
$2.10\times$. The failure does not belong to Darcy either. The
sixth family, a one-dimensional Fourier neural operator on reaction--diffusion, under-predicts at
$0.735$ and $0.895$ on two of its four rates, and reaction--diffusion is a family this section listed as one the probe is accurate on.

We have now proposed a location for this failure twice and been refuted twice by the next
dataset, so we stop proposing one. The probe under-predicts on some family-architecture pairs,
and we cannot say in advance which. The earlier attribution of that failure to Darcy rather than
to the architecture, the dimension, or the pairing of the two is the limitation we stated one
paragraph ago, and this run withdraws it. The same run also asked whether DeepONet
would avoid under-predicting on the new data; it did, at 3 of 3, but its measured costs
there are $1.002$, $0.996$ and $0.998$, so compression is invisible for it on shallow water and
those three cells cannot discriminate any predictor. We count that bar as met on evidence that
cannot carry it.

\end{document}